%% file: main.tex
\documentclass[a4paper,fleqn]{cas-dc}
\usepackage{booktabs}
\usepackage{amsmath}
\usepackage{wrapfig}

\usepackage[accsupp]{axessibility}  
\usepackage{bm}
\usepackage{multirow}
\usepackage{tikz}
\usepackage{comment}
\usepackage{amsmath,amssymb} 
\usepackage{color}
\usepackage{pifont}
\usepackage{makecell}
\usepackage{caption}
\usepackage{float}
\usepackage{adjustbox}
\usepackage{etoolbox}
\usepackage{dsfont}

\newcommand{\YKL}[1]{\textcolor{red}{#1}}

\usepackage[accsupp]{axessibility}  
\usepackage{paralist}

\def\eg{\textit{e.g.}} 
\def\ie{\textit{i.e}.} 

\def\etal{\textit{et al}.}

\definecolor{rblue}{rgb}{0,0.5,1}
\newcommand{\PAR}[1]{\noindent{\bf #1}}

\newcommand{\JQ}[1]{\textcolor{blue}{#1}}
\newcommand{\add}[1]{{\textcolor{black}{#1}}}

\usepackage{amsmath,lipsum}
\usepackage{cuted}
\usepackage{lineno}
\usepackage{booktabs,makecell,multirow}
\usepackage{amsmath,amssymb,amsfonts}
\usepackage{algorithmic}
\usepackage{graphicx}
\usepackage{textcomp}
\usepackage{amsmath,amsfonts}
\usepackage{algorithmic}
\usepackage{algorithm}
\usepackage{array}
\usepackage[caption=false,font=normalsize,labelfont=sf,textfont=sf]{subfig}
\usepackage{textcomp}
\usepackage{stfloats}
\usepackage{url}
\usepackage{verbatim}
\usepackage{graphicx}
\usepackage{cite}
\usepackage{booktabs}
\usepackage{paralist}

\usepackage{threeparttable}

\usepackage[numbers,sort&compress]{natbib}

\def\tsc#1{\csdef{#1}{\textsc{\lowercase{#1}}\xspace}}
\tsc{WGM}
\tsc{QE}

\usepackage{xspace}
\makeatletter
\DeclareRobustCommand\onedot{\futurelet\@let@token\@onedot}
\def\@onedot{\ifx\@let@token.\else.\null\fi\xspace}
\def\eg{\textit{e.g}\onedot} 
\def\ie{\textit{i.e}\onedot}

\def\etal{\textit{et al}\onedot}
\makeatother

\usepackage{color}

\usepackage{xcolor}
\definecolor{hollywoodcerise}{rgb}{0.96, 0.0, 0.63}
\definecolor{lasallegreen}{rgb}{0.03, 0.47, 0.19}
\definecolor{hanpurple}{rgb}{0.32, 0.09, 0.98}
\definecolor{green(pigment)}{rgb}{0.0, 0.65, 0.31}

\hypersetup{colorlinks=true,linkcolor={red},citecolor={hanpurple},urlcolor={magenta}} 

\usepackage{caption}
\begin{document}
\let\WriteBookmarks\relax
\def\floatpagepagefraction{1}
\def\textpagefraction{.001}

\shorttitle{Representing Domain-Mixing Optical Degradation for Real-World Computational Aberration Correction via Vector Quantization}    

\shortauthors{Q. Jiang \textit{et al.}}  

\title [mode = title]{Representing Domain-Mixing Optical Degradation for Real-World Computational Aberration Correction via Vector Quantization}  



%

\author[1,3]{Qi Jiang}
\credit{Conceptualization of this study, Methodology, Software}
\author[1,3]{Zhonghua Yi}
\author[1,3]{Shaohua Gao}
\author[1]{Yao Gao}
\author[1]{Xiaolong Qian}
\author[1,3]{Hao Shi}
\author[1]{Lei Sun}
\author[4]{JinXing Niu}
\author[1,3]{Kaiwei Wang}[orcid=0000-0002-8272-3119]
\cormark[1]
\ead{wangkaiwei@zju.edu.cn}
\author[2]{Kailun Yang} [orcid=0000-0002-1090-667X]
\cormark[1]
\ead{kailun.yang@hnu.edu.cn}
\author[1]{Jian Bai}

\affiliation[a]{organization={State Key Laboratory of Extreme Photonics and Instrumentation, College of Optical Science and Engineering},
            addressline={Zhejiang University}, 
            city={Hangzhou},
            postcode={310027}, 
            country={China}}

\affiliation[b]{organization={National Engineering Research Center of Robot Visual Perception and Control Technology},
            addressline={Hunan University}, 
            city={Changsha},
            postcode={410082}, 
            country={China}}

\affiliation[c]{organization={Intelligent Optics $\&$ Photonics Research Center, Jiaxing Research Institute},
            addressline={Zhejiang University}, 
            city={Jiaxing},
            postcode={314031}, 
            country={China}}

\affiliation[d]{organization={School of Mechanical Engineering},
            addressline={North China University of Water Resources and Electric Power}, 
            city={Zhenzhou},
            postcode={450045}, 
            country={China}}

\cortext[1]{Corresponding author}

\begin{abstract}
Relying on paired synthetic data, existing learning-based Computational Aberration Correction (CAC) methods are confronted with the intricate and multifaceted synthetic-to-real domain gap, which leads to suboptimal performance in real-world applications. In this paper, in contrast to improving the simulation pipeline, we deliver a novel insight into real-world CAC from the perspective of Unsupervised Domain Adaptation (UDA). By incorporating readily accessible unpaired real-world data into training, we formalize the Domain Adaptive CAC (DACAC) task, and then introduce a comprehensive Real-world aberrated images (Realab) dataset to benchmark it. The setup task presents a formidable challenge due to the intricacy of understanding the target optical degradation domain. To this intent, we propose a novel Quantized Domain-Mixing Representation (QDMR) framework as a potent solution to the issue. Centering around representing and quantizing the optical degradation which is consistent across different images, QDMR adapts the CAC model to the target domain from three key aspects: (1) reconstructing aberrated images of both domains by a VQGAN to learn a Domain-Mixing Codebook (DMC) characterizing the optical degradation; (2) modulating the deep features in CAC model with DMC to transfer the target domain knowledge; and (3) leveraging the trained VQGAN to generate pseudo target aberrated images from the source ones for convincing target domain supervision. Extensive experiments on both synthetic and real-world benchmarks reveal that the models with QDMR consistently surpass the competitive methods in mitigating the synthetic-to-real gap, which produces visually pleasant real-world CAC results with fewer artifacts. Codes and datasets are made publicly available at \url{https://github.com/zju-jiangqi/QDMR}.
\end{abstract}

\maketitle

\section{Introduction}
\input{introduction_revised}

\section{Related Work}
\input{related_work}

\section{Benchmark Settings}
\input{problem}

\section{Methodology}
\input{method}
\section{Experiments}

\input{experiments}

\section{Conclusion and Discussion}
\input{conclusion}

\section*{Funding}
National Key R\&D Program of China, under Grant 2022YFF0705500. 
National Natural Science Foundation of China (No. 62473139). 
Henan Province Key R\&D Special Project (231111112700). 
Hangzhou SurImage Technology Company Ltd.

%
\bibliographystyle{SELF-cas-model2-names}
\bibliography{ref}

\input{supp}

\end{document}

%% file: introduction_revised.tex
Computational Aberration Correction (CAC)~\cite{Schuler2012Blind,2013High,peng2019learned,chen2021extreme_quality,eboli2022fast}, in which a post-processing model is applied to deal with the degradation (\ie, the optical degradation~\cite{Chenshiqi}) induced by residual optical aberrations of the target optical system, delivers a viable solution for the applications of compact and lightweight optical systems with limited optical elements. 
Driven by the capacity of deep neural networks,
the emerging learning-based methods~\cite{Chenshiqi,chen_mobile_2023,jiang2023minimalist} have significantly improved the results of CAC, which also formulates the task of supervised learning on paired images.
Confronted with the challenges of capturing clear-aberrated image pairs in real-world scenes, imaging simulation~\cite{Chenshiqi,sun2021end} has become a popular method to generate synthetic data in mass, \add{facilitating the supervised training of CAC models.}

\begin{figure}[!t]
  \centering
  \includegraphics[width=1.0\linewidth]{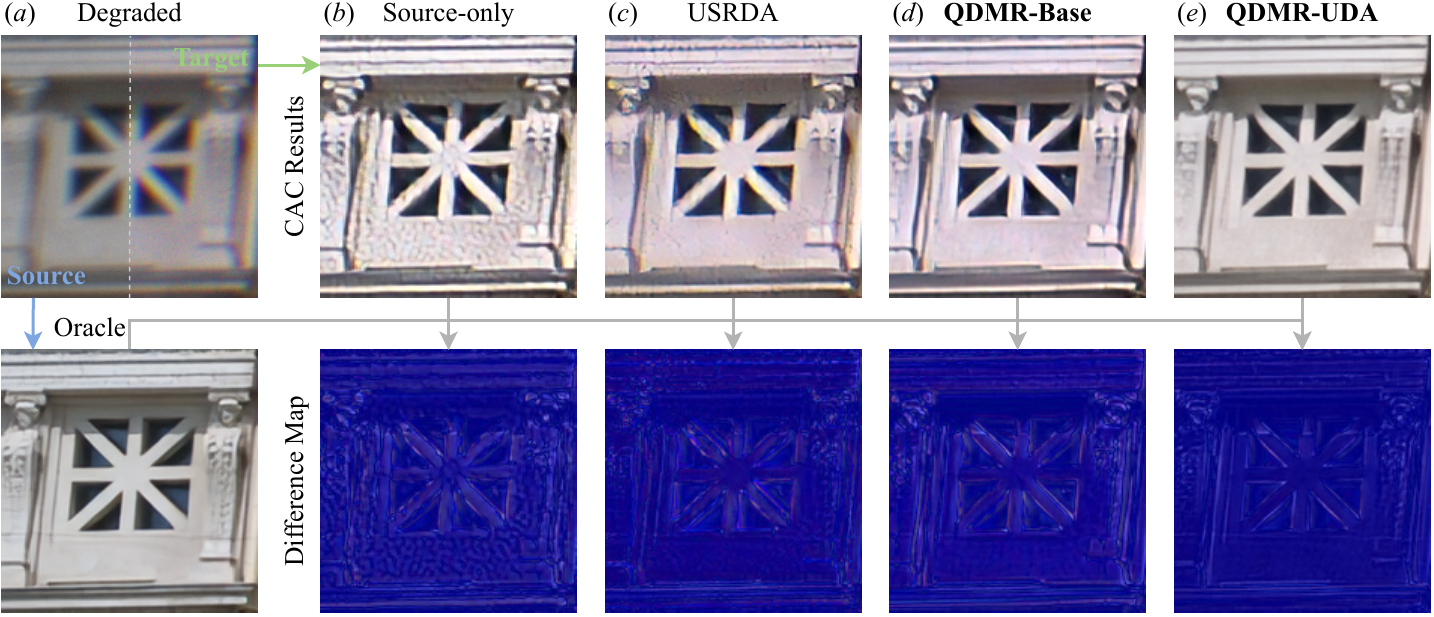}
  \caption{\textbf{Visualization of the synthetic-to-real gap problem in real-world CAC.} The proposed QDMR methods present a powerful solution to the issue. The ``Oracle'' and ``Source-only'' here denote that the model trained on the source images is tested on the source and target images respectively. (a): The domain gap between the synthetic aberrated image (source) and the real-world one (target). (b)-(e): The CAC results and difference maps (absolute difference between CAC results and oracle performance) of the same model under different training pipelines.  }
  \label{fig:intro}
\end{figure}

\add{However, the current CAC models trained on synthetic data can hardly deliver comparable results in real-word aberrated images as synthetic ones, suffering from severe artifacts, as shown in Figure~\ref{fig:intro} (a) and (b).} 
\add{This is due to the \textbf{non-negligible domain gap between synthetic data and real-world scenes}, where the simulation pipeline cannot accurately simulate the real imaging situations.}
Meanwhile, the factors that cause the domain gap are diverse and complex~\cite{lin2023learning}, \eg, aberration deviation due to the errors in the simulation model and manufacture-assembly process of the lens, the difference in Field of View (FoV) between ground-truth images and real-world images, different shooting distances, and various types of imaging sensors.
Consequently, even with the strategies for mitigating the gap, such as calibration and disturbance~\cite{chen2022computational_mass,hu2021image,jiang2023annular}, the current pipelines struggle to accurately simulate real-world aberrated images, bringing limitations to downstream real-world applications of CAC methods~\cite{jiang2023minimalist, jiang2022computational}.

\add{Recent blooming development of \textit{Unsupervised Domain Adaptation (UDA)}~\cite{hsu2020progressive, hoyer2022daformer} provides a novel insight into this issue, which aims to adapt the model from a labeled source domain to an unlabeled target domain for closing the domain gap.}
\add{In our real-world CAC, the paired synthetic aberrated images can be treated as the labeled source domain, while the real-world ones without ground truth can be considered as the unlabeled target domain. }
\add{In this way, instead of improving the optical simulation model, we intend to develop a UDA solution to directly train the model on real-world images with the assistance of synthetic images, formalizing the task of \textit{Domain Adaptive Computational Aberration Correction (DACAC).}}
To facilitate this task, we establish a comprehensive benchmark, where a new dataset \textit{RealAb} is proposed composed of paired synthetic aberrated images \textit{Syn}, and real-world aberrated images by snapping (\textit{Real-Snap}) and domain gap simulation (\textit{Real-Sim}), providing both quantitative and qualitative evaluation protocols.

Despite that advanced UDA methods are emerging in high-level perception tasks (\eg, object detection~\cite{hsu2020progressive,mattolin2023confmix,li2023domain} and semantic segmentation~\cite{hoyer2022hrda,hoyer2022daformer,jia2023dginstyle}), seldom explorations have been made in low-level vision, which is often task-specific~\cite{shao2020domain,chen2021psd}. 
The most common designs build on adversarial domain data transformation and feature alignment, but they can hardly achieve promising results in DACAC, \eg~USRDA~\cite{wang2021unsupervised} (see Figure~\ref{fig:intro} (c)), due to the difficulty of learning the multifaceted domain shift. 
In this case, a more effective framework is needed for solving the challenging DACAC task.

Considering that the optical degradation of an optical system is consistent across different images, the key is to understand the optical degradation of the applied optical system in the target domain.
As Feynman once said, \textit{``What I cannot create, I do not understand''}. In this paper, we propose to mix and reconstruct aberrated images (\textit{create}) of both domains via a VQGAN~\cite{van2017neural, esser2021taming}, to learn the Quantized Domain-Mixing Representation (QDMR) for \textit{understanding} the domain gap. 
During the Vector-Quantized (VQ) codebook learning stage, the essential information pertaining to the reconstruction, \ie, the domain-mixing priors of optical degradation, is characterized and learned within a trainable codebook, which is coined Domain-Mixing Codebook (DMC).
In this way, we design the QDMR-Base model for DACAC with the guidance of DMC, which is leveraged to modulate the image restoration features through feature quantization and affine transformation.
The experimental results in Figure~\ref{fig:intro} (d) reveal that the QDMR-Base can deliver impressive restored images with sharper edges and fewer artifacts, illustrating that the DMC successfully transfers the target domain knowledge to the CAC model.  

Moreover, we find that the remarkable capacity of the trained VQGAN in QDMR to generate target real-world aberrated images showcases potential in the \textit{source domain to the target domain Transformation (s2tT)}.
In this case, an s2tT constraint is devised to refine the pretraining objective, which enables the VQGAN to transform source synthetic images to target real-world images.
We propose the \textit{QDMR-UDA} framework, which utilizes the supervision of pseudo target image pairs from s2tT and common applied UDA strategy, \ie, adversarial domain Feature Alignment (FA), to further adapt the base model to the target domain. 
As illustrated in Figure~\ref{fig:intro} (e), compared to the competitive UDA framework~\cite{wang2021unsupervised} and QDMR-Base, \textbf{only QDMR-UDA can effectively mitigate the issue of domain gap in CAC, generating realistic aberration-free image comparable to the oracle result. }

To the best knowledge of the authors, this is the first work formulating the real-world computational aberration correction as an unsupervised domain adaptation task. The main contributions are summarized as follows:
\begin{compactitem}
    \item We deliver a novel insight into the issue of the synthetic-to-real gap in CAC from the perspective of UDA and formalize the task of DACAC, where a real-world aberrated image dataset RealAb is put forward. 
    \item We propose the QDMR to learn the mixing optical degradation priors of aberrated images on both domains by a VQGAN, in a self-supervised way, which guides the restoration stage in the QDMR-Base model to produce domain adaptive CAC results.
    \item We introduce the QDMR-UDA framework equipped with UDA strategies, where the s2tT. ability of the trained VQGAN and the FA strategy are explored to further adapt the model to the target domain.
    \item Our proposed QDMR method offers a superior solution to the issue of domain gap in CAC, surpassing all competitive CAC models and UDA frameworks on both synthetic and real-world benchmarks.  
\end{compactitem}

%% file: related_work.tex
\noindent\textbf{Computational Aberration Correction.}
Due to insufficient optical lenses for aberration correction, the imaging results of compact and light-weight optical systems suffer from optical degradation~\cite{2011Modeling,mahajan1994zernike}. 
Serving as a post-image-processing model, the Computational Aberration Correction (CAC) method~\cite{2011Non,2013High} is proposed to enhance the image quality of these systems, which is a classical computational imaging framework.
Early efforts have been made to solve the inverse problem through model-based methods~\cite{Schuler2012Blind,2015Blind}.
Recently, learning-based methods~\cite{peng2019learned,chen2021extreme_quality,chen_mobile_2023} have been widely explored for delivering more impressive results in CAC, which benefits from the blooming development of image restoration~\cite{liang2021swinir,chen2022simple}, image Super-Resolution (SR)~\cite{wang2018esrgan,zhang2018image,liang2021swinir} and image Deblur~\cite{zamir2022restormer,chen2022simple,wang2022uformer} methods.

However, these methods create a significant demand for paired training data, which is often the synthetic data produced by optical simulation~\cite{Chenshiqi}. 
In this case, the synthetic-to-real domain gap comes to the forefront, where the optical model cannot accurately simulate the real imaging situations, preventing promising results in real-world scenes. 
Several works have focused on improving the accuracy of optical models~\cite{Chenshiqi}, calibrating the manufactured lens~\cite{chen2022computational_mass}, or adding disturbance augmentation~\cite{hu2021image,jiang2023annular}, attempting to mitigate the gap. 
Yet, the domain gap always exists due to inevitable approximation in simulation~\cite{lin1994modeling}, different FoVs of ground-truth~\cite{gu2021pit}, different shooting distances~\cite{liu2021end,luo2024correcting}, and different sensors~\cite{brooks2019unprocessing}. 
In this paper, we make the first attempt to formalize the task of Domain Adaptive Computational Aberration Correction (DACAC), solving the issue of the synthetic-to-real gap from the perspective of Unsupervised Domain Adaptation (UDA). 
The unpaired real-world data are incorporated into the training process to enhance the domain generalization ability of the CAC model.

\noindent\textbf{Domain Adaptation in Low-Level Vision.}
The domain adaptation aims to adapt the model from a labeled source domain to an unlabeled target domain, which has been explored to address the issue of the synthetic-to-real gap in some low-level vision tasks, \eg, image super-resolution~\cite{wei2021unsupervised,wang2021unsupervised}, image dehazing~\cite{li2019semi,chen2021psd,li2023dadrnet}, \add{image deraining~\cite{yasarla2021semi,chang2023unsupervised},} and underwater image enhancement~\cite{jiang2022two,wang2023domain}. 
However, \add{these frameworks often utilize task-specific physical priors (\eg, dark channel prior)~\cite{chen2021psd,li2019semi}, physical properties (\eg, decoupling)}~\cite{chang2023unsupervised, li2023dadrnet}, or additional modalities (\eg, depth)~\cite{shao2020domain}, which can hardly be adopted in our DACAC task. 
The most common universal strategies are adversarial domain feature alignment~\cite{wei2021unsupervised,wang2023domain,li2023dadrnet} and domain data transformation~\cite{guo2020closed,wang2021unsupervised,jiang2022two}, which enables the model to learn the domain-invariant feature extraction, delivering valuable solutions for DACAC. 
\add{To fill the research gap of UDA methods in DACAC, based on the cross-scene-consistent property of optical degradation, a self-supervised method is proposed to learn its domain-mixing representation, serving as a powerful baseline for the task.} 
We initially assess the effectiveness of the above UDA strategies atop the baseline, designing a UDA framework for unlocking the potential of the representation to achieve better DACAC results.

\noindent\textbf{Vector-Quantized Codebook Learning.}
Vector-Quantized (VQ) codebook learning~\cite{van2017neural,esser2021taming,zhang2023regularized} reveals strong ability to represent domain-invariant prior of natural images with entries of a codebook.
The learned codebook is then explored to improve the performance of down-stream low-level vision tasks, \eg, image super-resolution~\cite{chen2022real}, image dehazing~\cite{wu2023ridcp}, face restoration~\cite{gu2022vqfr,zhou2022towards}, \add{and adverse weather removal~\cite{ye2023adverse}, where the features of the low-quality image are matched with the High-Quality Prior (HQP) in the codebook to reconstruct the high-quality image.}
Nevertheless, the matching process is still domain-specific, where the key feature matching has not been trained on target real-world data, leading to the domain gap problem.
Meanwhile, Chen~\etal~\cite{chen_mobile_2023} show that learning image degradation priors via codebook can also boost the performance of the CAC model. 
\add{To this intent, we propose to explore VQ codebook learning for aberrated images rather than clear images~\cite{wu2023ridcp,ye2023adverse}, where the target images can be incorporated into the training stage, to learn a Quantized Domain-Mixing Representation (QDMR) to bridge the domain gap.}
The learned Domain-Mixing Codebook (DMC) is utilized to guide the image restoration and further adapt the model to the target domain, delivering outstanding performance in DACAC tasks.

%% file: problem.tex
To deal with the domain gap issue in real-world Computational Aberration Correction, we first formalize it as an Unsupervised Domain Adaptation (UDA) problem, \ie, Domain Adaptive Computational Aberration Correction (DACAC), in Section~\ref{subsec:dacac}. 
Then, the datasets are introduced and based on which the benchmarks for DACAC are set up in Section~\ref{subsec:benchmark}. 
In this way, we pioneer addressing the synthetic-to-real domain gap from a new perspective, which focuses on the domain adaptability of the CAC model.

\subsection{The DACAC Task}
\label{subsec:dacac}
The traditional learning-based CAC task often generates synthetic aberrted images $\mathcal{Y}_{S}{=}\{{y_S}^{(i)}\}_{i=1}^{N}$ from the real-world clear images $\mathcal{X}{=}\{{x}^{(i)}\}_{i=1}^{N}$ via an optical simulation model~\cite{Chenshiqi} for model training . The trained model is then applied to the real-world aberrated images $\mathcal{Y}_{R}{=}\{{y_R}^{(i)}\}_{i=1}^{M}$, which suffers from the domain gap between $\mathcal{Y}_{S}$ and $\mathcal{Y}_{R}$, achieving unpromising results. 

In contrast to improving the simulation model, we propose to focus on enhancing the domain generalization ability of the CAC model, by incorporating the unpaired real-world data $\mathcal{Y}_{R}$ into the training process, \ie, formalizing the task of Domain Adaptive Computational Aberration Correction (DACAC).
Specifically, given the paired synthetic data, which is defined as the source domain data: $\mathcal{D}_{S}{=}\{{y_S}^{(i)}, {x}^{(i)}\}_{i=1}^{N}$, and the unpaired real-world data, which is defined as the target domain: $\mathcal{D}_{T}{=}\{{y_R}^{(i)}\}_{i=1}^{M}$, the DACAC task aims to learn a CAC model on both $\mathcal{D}_{S}$ and $\mathcal{D}_{T}$ to predict accurate clear images from unseen real-world aberrated images. 

\begin{figure}[!h]
  \includegraphics[width=1.0\linewidth]{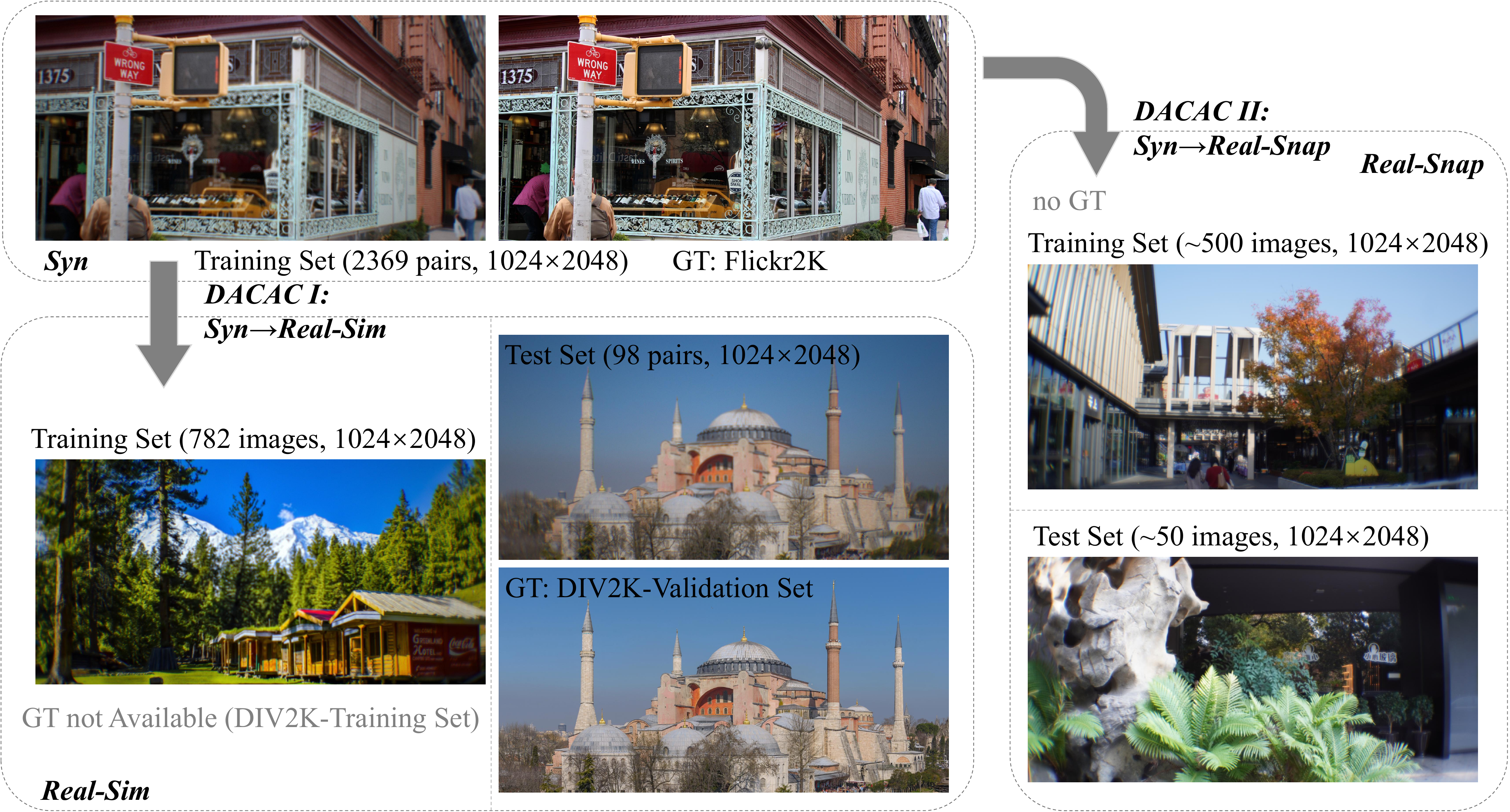}
  \caption{\add{\textbf{Overview of the established Realab dataset and corresponding DACAC tasks.} We take the dataset for MOS-S1 as an example. The paired source domain training data \textit{Syn} is generated based on Flickr2K~\cite{timofte2017ntire}, while the target domain training data \textit{Real-Sim} is generated based on DIV2K~\cite{timofte2017ntire}, where the ground-truth is not available during training. For a comprehensive evaluation, the paired test data in \textit{Real-Sim} can provide numerical evaluation under convincing referenced metrics, and the test images in \textit{Real-Snap}, captured by a real lens, are leveraged to deliver intuitive real-world CAC performance.}}
  \label{fig:benchmark}
\end{figure}

\subsection{Datasets and Benchmarks}
\label{subsec:benchmark}
To facilitate the relative research, we benchmark the DACAC task with a novel Real-world aberrated images dataset (\textit{Realab}). Realab is composed of aberrated images under two Minimalist Optical Systems (MOS) with distinct aberration behaviors (MOS-S1 and MOS-S2), which is a typical application of CAC~\cite{2013High,jiang2023minimalist}. 
For $\mathcal{D}_{S}$, we directly feed the ray-tracing-based optical model~\cite{Chenshiqi} with the design parameters of the applied MOS to generate paired images, \ie, coined \textit{Syn}.
For $\mathcal{D}_{T}$, two settings are applied:
(1) \textit{Real-Snap}, where we snap real-world images with the two MOS; (2) \textit{Real-Sim}, where we add disturbance to the parameters of the optical model to simulate the synthetic-to-real gap and generate paired ``real-world'' images, where the ground truth is only available for the test set.
In this regard, as shown in Figure~\ref{fig:benchmark}, the DACAC task is performed on \textbf{\textit{Syn$\to$Real-Snap}} and \textbf{\textit{Syn$\to$Real-Sim}}, respectively. Comprehensively, the former results showcase the intuitive real-world performance, while the latter results provide numerical evaluation under referenced metrics, \eg~LPIPS~\cite{zhang2018unreasonable}, FID~\cite{heusel2017gans}, PSNR, and SSIM~\cite{wang2004image}.
\add{Please refer to Section~\ref{subsec:dataprepare} for more details regarding the dataset preparation.}

%% file: method.tex
\begin{figure*}[!t]
  \centering
  \includegraphics[width=1.0\linewidth]{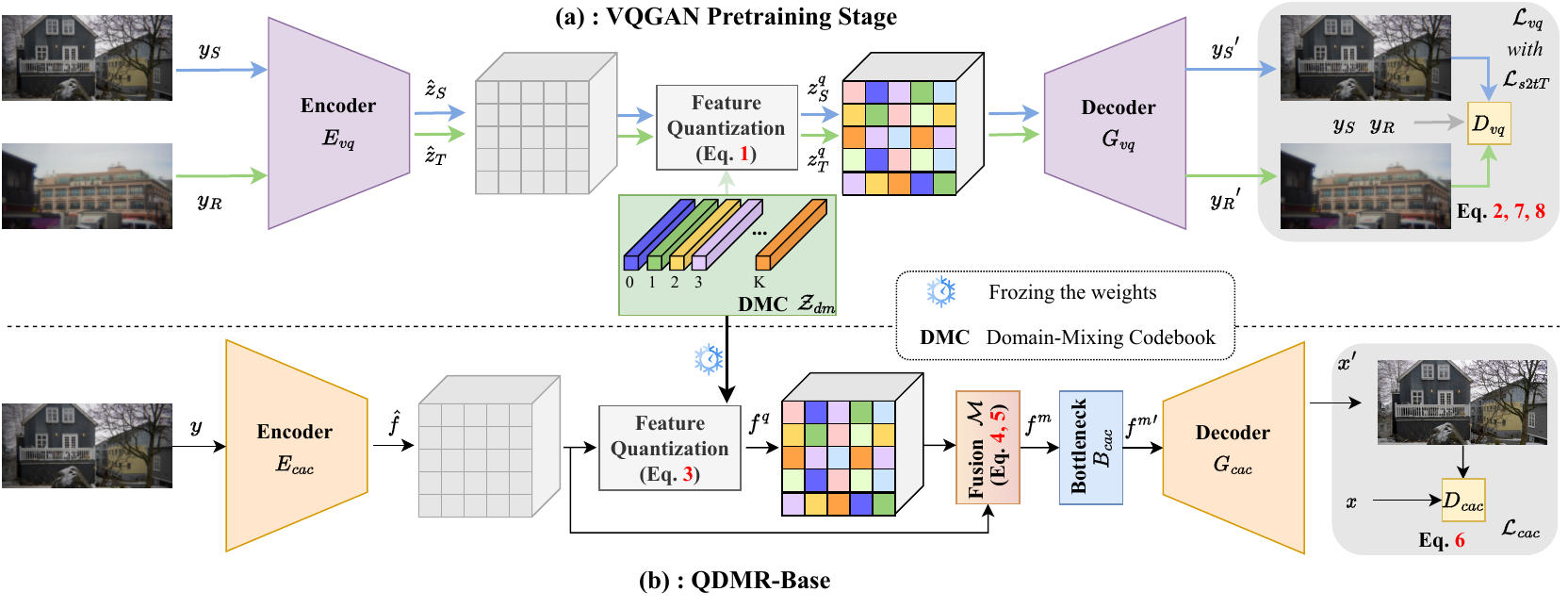}
  \caption{\textbf{Overview of the proposed QDMR.} (a): To characterize the domain-mixing priors of optical degradation, we first pretrain a VQGAN to learn the DMC. (b): The pre-trained DMC is leveraged to guide the image restoration feature in the QDMR-Base model. The bottleneck module $B_{cac}$ can be any backbone for low-level vision tasks. }
  \label{fig:qdmr}
\end{figure*}

The key idea of our method is to learn the domain-mixing image degradation priors of aberrated images via self-supervised VQ codebook learning. 
The overall framework is illustrated in Figure~\ref{fig:qdmr}. 
In Section~\ref{subsec:qdmr}, we first adopt the VQGAN~\cite{esser2021taming} to reconstruct aberrated images on both domains to learn a Quantized Domain-Mixing Representation (QDMR), which is then explored to guide the CAC model to achieve domain adaptive results in Section~\ref{subsec:qdmrbase}.
To further adapt the model to the target domain, we exploit the VQGAN's generation ability of target images, and propose the QDMR-UDA framework in Section~\ref{subsec:qdmruda}.   

\subsection{Quantized Domain-Mixing Representation}
\label{subsec:qdmr}
In VQGAN, given a codebook $\mathcal{Z}=\{z_k\}_{k=1}^{K}\in\mathbb{R}^{n}$, the latent feature  $\hat{z}\in\mathbb{R}^{h{\times}w{\times}n}$ of the input image with spatial size $(h\times{w})$ and channel dimension $n$, is quantized by finding the nearest neighbours in $\mathcal{Z}$ for its each element $\hat{z}_{ij}$, to calculate the discrete representation ${z}^{q}\in\mathbb{R}^{h{\times}w{\times}n}$:
\begin{equation}
\label{eq:codematching}
{z}^{q}_{ij} = \arg\min\limits_{z_k\in\mathcal{Z}}(\Vert{\hat{z}_{ij}-z_k}\Vert_2),
\end{equation}
where $K$ denotes the codebook size and $i\in\{1,2,\cdots,h\}$, $j\in\{1,2,\cdots,w\}$ denote the coordinates in the feature space. 
The quantized feature ${z}^{q}$ is then applied to reconstruct the input image for self-supervised learning.
After training VQGAN, the codebook represents the priors for the input image domain, which delivers potential in guiding the downstream image restoration tasks~\cite{gu2022vqfr,zhou2022towards}. 

To understand how the image degrades over different domains, especially the unseen target domain, we propose to learn the Quantized Domain-Mixing Representation (QDMR), where both the source and target aberrated images are reconstructed by the VQGAN. 
Specifically, as illustrated in Figure~\ref{fig:qdmr} (a), the latent features $\{\hat{z}_S, \hat{z}_T\}$ of source and target aberrated input $\{y_{S}, y_{R}\}$ extracted by a shared encoder ${E}_{vq}$, are quantized by a learnable Domain-Mixing Codebook (DMC) $\mathcal{Z}_M$, to obtain the quantized features $\{{z}^q_S, {z}^q_T\}$. 
$\{{z}^q_S, {z}^q_T\}$ are then fed into a shared decoder ${G}_{vq}$ to reconstruct the source and target aberrated images $\{{y}'_S, {y}'_R\}$. 
Following~\cite{chen2022real}, the VQGAN is trained via the following objective function: 
\begin{equation}
\label{eq:vqganloss_s}
\mathcal{L}_{vq} = {\Vert{y'_{S(R)}-y_{S(R)}}\Vert_1} + \lambda_{per}^{vq}\mathcal{L}_{per} + \lambda_{adv}^{vq}\mathcal{L}_{adv} + \mathcal{L}_{codebook} ,
\end{equation}
where $\mathcal{L}_{per}$ is the perceptual loss, $\mathcal{L}_{adv}$ is the adversarial loss, and $\mathcal{L}_{codebook}$ is the codebook loss to optimize $\mathcal{Z}_M$. 
It is worth mentioning that $\mathcal{L}_{adv}$ in two domains share the same discriminator ${D}_{vq}$, which enables the VQGAN to generate the domain-mixing aberrated images via adversarial training.
In this way, the learned DMC $\mathcal{Z}_{dm}$ implicitly indicates the degradation-aware priors of both domains, which is explored to guide the CAC model to adapt to the target domain in the next subsection.   

\begin{figure*}[!h]
  \centering
  \includegraphics[width=0.8\linewidth]{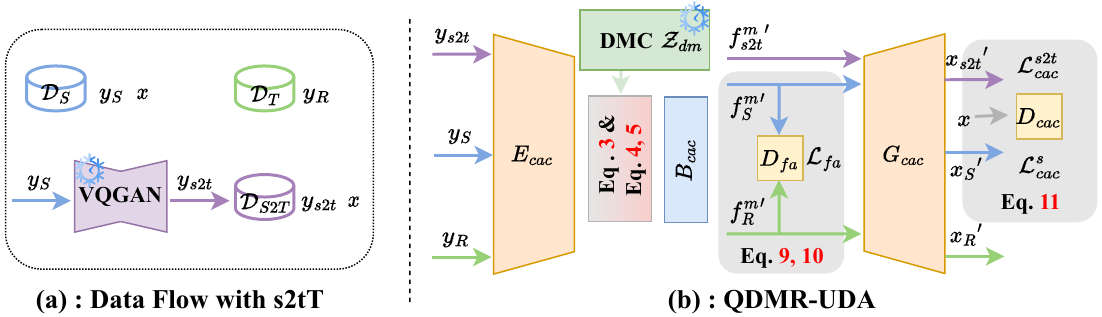}
  \caption{\textbf{Illustration of the proposed QDMR-UDA framework.} (a): The VQGAN is exploited to transform the source images to the target domain, which generates pseudo paired target images for supervision. (b): Based on the s2t data flow and the QDMR-Base model, we develop the QDMR-UDA framework to further adapt the base model to the target domain through UDA training with a feature alignment strategy.}
  \label{fig:qdmruda}
\end{figure*}

\subsection{QDMR-based Computational Aberration Correction}
\label{subsec:qdmrbase}
To transfer the learned priors to the CAC model, we propose the QDMR-Base model by modulating the restoration feature with DMC. 
As shown in Figure~\ref{fig:qdmr} (b), a U-Net CAC model~\cite{chen2022real} is first established as the baseline, where the encoder ${E}_{cac}$ extracts deep feature $\hat{f}$ from the aberrated input $y$, and the decoder ${G}_{cac}$ restores the clear image $x$ from the processed feature $\hat{f'}={B}_{cac}(\hat{f})$ of the bottleneck module ${B}_{cac}$. 
To distill the priors in DMC, we quantize the deep feature $\hat{f}$ by $\mathcal{Z}_{dm}$ similar to Eq.~\ref{eq:codematching}:
\begin{equation}
\label{eq:codematchingcac}
{f}^{q}_{ij} = \arg\min\limits_{z_k\in\mathcal{Z}_{dm}}(\Vert{\hat{f}_{ij}-z_k}\Vert_2),
\end{equation}
and then obtain the QDMR feature ${f}^q$ which indicates the degradation information of the input aberrated image. 
However, it is challenging to directly predict the clear image from ${f}^q$ due to the following two reasons:
(1) the DMC contains no priors of the clear domain; (2) the quantization process leads to information loss.  
To this intent, we propose a fusion module to utilize the QDMR feature ${f}^q$ for modulating the origin deep feature $\hat{f}$, which not only preserves the ability of the CAC model to restore aberrated images but also helps the model to understand the target domain from QDMR. 
Concretely, as a common practice in feature fusion~\cite{Wang_2018_CVPR,li2022all,ai2023multimodal,zhang2023efficient}, a mapping function $\mathcal{M}$ is learned from ${f}^q$ to conduct affine transformation on $\hat{f}$ by scaling and shifting with the predicted modulation parameters $(\gamma, \beta)$:
\begin{equation}
\label{eq:m}
(\gamma, \beta) = \mathcal{M}({f}^q),
\end{equation}
\begin{equation}
\label{eq:sft}
{f}^m = \gamma\odot{\hat{f}} + \beta,
\end{equation}
where $f^m$ is the modulated feature and the mapping function $\mathcal{M}$ is implemented by two $1{\times}1$ convolution layers. 
Finally, the modulated feature $f^m$ is further processed by ${B}_{cac}$ to obtain the restoration feature ${f^m}'$, which is fed to the ${G}_{cac}$ for predicting the aberration-free image $x'$.
We follow~\cite{wu2023ridcp} to optimize the QDMR-Base model with L1, perceptual, and adversarial losses for generating visual-pleasant and realistic CAC results. The training objective is written as:
\begin{equation}
\label{eq:cacloss}
\mathcal{L}_{cac} = {\Vert{x'-x}\Vert_1} + \lambda_{per}^{cac}\mathcal{L}_{per} + \lambda_{adv}^{cac}\mathcal{L}_{adv}.
\end{equation}
Considering that $\mathcal{L}_{cac}$ is a supervised loss function, we only optimize the QDMR-Base model on the source domain data $\mathcal{D}_{S}{=}\{{y_S}^{(i)}, {x}^{(i)}\}_{i=1}^{N}$. 
Note that the DMC $\mathcal{Z}_{dm}$ is frozen during the training process to preserve the learned knowledge of the target domain for guiding the CAC model.

\subsection{QDMR-based Unsupervised Domain Adaptation}
\label{subsec:qdmruda}
The QDMR-Base model serves as a powerful baseline for the DACAC task, based on which we further propose the QDMR-UDA framework to improve the training of its restoration stage. 
The core design is to incorporate the target domain data into the training. 
As illustrated in Figure~\ref{fig:qdmruda}, we make the following efforts:

\PAR{Data Transformation from Source to Target.}
Benefiting from the shared discriminator $D_{vq}$, the reconstructed aberrated images of VQGAN in both source and target domains will be put close to each other, which means the input source data can be transformed into the target one. 
Motivated by the observation, we utilize the trained VQGAN to apply the source domain to the target domain Transformation (s2tT).
To be specific, the source image ${y}_S$ are fed into the VQGAN to produce the s2tT result ${y}_{s2t}$, which constructs the paired data $\{\mathcal{Y}_{s2t}, \mathcal{X}\}$ to enable supervised training in the target domain. 
Intuitively, the domain adaptive results are greatly affected by $y_{s2t}$, so we further put forward a s2tT constraint $\mathcal{L}_{s2tT}$ to replace the $\mathcal{L}_{adv}$ in Eq.~\ref{eq:vqganloss_s}.
The shared $D_{vq}$ is trained to discriminate real target image $y_R$ from the generated ones $\{{y_R}', {y_S}'\}$, while VQGAN is expected to fool the $D_{vq}$. We formulated the constraint as:
\begin{equation}
\label{eq:s2t_G}
\begin{split}
&\mathcal{L}_{s2tT}(E_{vq}, \mathcal{Z}_{dm}, G_{vq}) = \mathbb{E}_{\mathcal{D}_{S}}(D_{vq}(y_{S}')-1)^2 \\
&+ \mathbb{E}_{\mathcal{D}_{T}}(D_{vq}(y_{R}')-1)^2,
\end{split}
\end{equation}

\begin{equation}
\label{eq:s2t_D}
\begin{split}
&\mathcal{L}_{s2tT}(D_{vq}) = \mathbb{E}_{\mathcal{D}_{S}}(D_{vq}(y_{S}')-0)^2 \\
&+  \mathbb{E}_{\mathcal{D}_{T}}(D_{vq}(y_{R}')-0)^2 \\
&+ \mathbb{E}_{\mathcal{D}_{T}}(D_{vq}(y_{R})-1)^2,
\end{split}
\end{equation}
where the form of LSGAN~\cite{mao2017least} is employed for the loss function to stabilize the training. With $\mathcal{L}_{s2tT}$, the VQGAN can not only learn the DMC for guiding the restoration stage but also provide the convincing pseudo paired target data $\mathcal{D}_{S2T}= \{\mathcal{Y}_{s2t}, \mathcal{X}\}$ without training another data transformation model. 

\PAR{Adversarial Domain Feature Alignment.}
Furthermore, we introduce a feature-level adversarial loss $\mathcal{L}_{fa}$ to supervise the $E_{cac}$, $\mathcal{M}$, and $B_{cac}$ to produce domain-invariant feature.
In this way, the $G_{cac}$ trained on source data can predict the accurate clear image from the aligned target feature, whose distribution is constrained close to that of the source domain.
With a feature-level discriminator $D_{fa}$, $\mathcal{L}_{fa}$ is also defined based on LSGAN~\cite{mao2017least} loss:
\begin{equation}
\label{eq:fa_G}
\begin{split}
&\mathcal{L}_{fa}(E_{cac}, \mathcal{M}, B_{cac}) = \lambda_{s}\mathbb{E}_{\mathcal{D}_{S}}(D_{fa}({f_{S}^m}')-0.5)^2\\  
&+ \lambda_{t}\mathbb{E}_{\mathcal{D}_{T}}(D_{fa}({f_{R}^m}')-0.5)^2,
\end{split}
\end{equation}

\begin{equation}
\label{eq:fa_D}
\begin{split}
&\mathcal{L}_{fa}(D_{fa}) = \mathbb{E}_{\mathcal{D}_{T}}(D_{fa}({f_{R}^m}')-0)^2\\ 
&+ \mathbb{E}_{\mathcal{D}_{S}}(D_{fa}({f_{S}^m}')-1)^2,
\end{split}
\end{equation}
where ${f_{S}^m}'$ and ${f_{R}^m}'$ are restoration features of the source and target domain before the decoder, and $(\lambda_{s}, \lambda_{t})$ are loss weights for two domains. 

\PAR{Training Objectives.}
In QDMR-UDA, we apply three data flows to train the CAC model based on $\mathcal{D}_{S}$, $\mathcal{D}_{S2T}$, and $\mathcal{D}_{T}$, as shown in Figure~\ref{fig:qdmr} (c).
The overall training objective of QDMR-UDA is: 
\begin{equation}
\label{eq:total}
\mathcal{L}_{uda} = \lambda_{s}\mathcal{L}_{cac}^{s} + \lambda_{s2t}\mathcal{L}_{cac}^{s2t} + \lambda_{fa}\mathcal{L}_{fa}.
\end{equation}
$\mathcal{L}_{cac}^{s}$ and $\mathcal{L}_{cac}^{s2t}$ adopt the supervised training loss in Eq.~\ref{eq:cacloss}, whose adversarial losses share the same discriminator ${D}_{cac}$ to produce domain-invariant restoration results, for paired data $\mathcal{D}_{S}$ and $\mathcal{D}_{S2T}$, respectively.
The unpaired data $\mathcal{D}_{T}$ is also incorporated into the training via the feature alignment in Eq.~\ref{eq:fa_G} with $\mathcal{D}_{S}$. 

\add{The implementation details of the QDMR framework will be depicted in Section.~\ref{subsec:details}.}

%% file: experiments.tex
\subsection{\add{Dataset Preparation}}
\label{subsec:dataprepare}
\add{As shown in Figure~\ref{fig:benchmark}, our dataset, \ie, Realab, consists of three components: \textit{Syn}, \textit{Real-Sim}, and \textit{Real-Snap}, containing data for training and evaluation of the DACAC task under our two applied optical systems. In this section, we will introduce the detailed creation process and data characteristics of Realab.}

\begin{figure}[!h]
  \centering
  \includegraphics[width=1.0\linewidth]{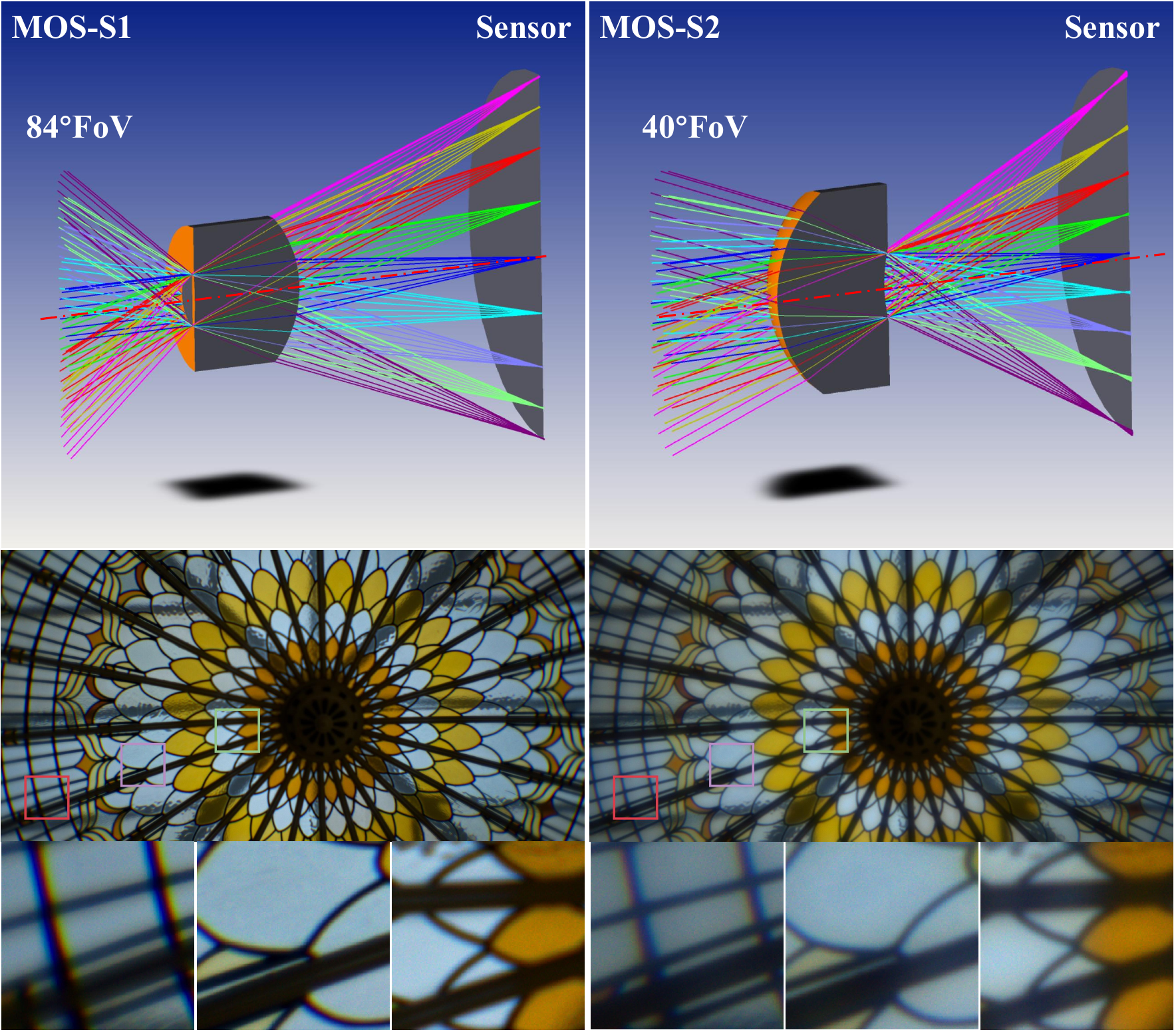}
  \caption{\textbf{The applied optical systems in Realab.} First row: the optical paths of the applied MOS. Bottom two rows: Imaging results of the two optical systems with different aberration behaviors. }
  \label{fig:opsystem}
\end{figure}

\PAR{Applied Optical Systems.}
As shown in the first row of Figure~\ref{fig:opsystem}, the two applied MOS are both composed of a single spherical lens, but their surface types differ in concavity, which results in different behaviors of aberrations (coined MOS-S1 and MOS-S2).
To be specific, the optical degradation of MOS-S1 reveals spatially variant distributions over FoVs, whereas that of MOS-S2 shows uniform distributions with severer aberrations (see bottom two rows of Figure~\ref{fig:opsystem}).  
The maximum half FoV is $42^\circ/20^\circ$, and the average RMS spot radius is $79.18{\mu}m/79.59{\mu}m$ for MOS-S1/S2 respectively. 
We attempt to establish a convincing benchmark by evaluating the combined performance under these two distinct aberration behaviors.
In addition, we apply the \textit{Sony $\alpha6600$} camera with the pixel size of $3.9 {\mu}m$ to snap real-world aberrated images equipped with the two MOS, whose Image Signal Processing (ISP) pipeline will be considered in the generation of \textit{Real-Sim} data. 

\PAR{Generation of \textit{Syn}.}
Based on the ray-tracing-based simulation model~\cite{Chenshiqi}, we generate the paired synthetic aberrated images for the source domain data in the DACAC task.
Concretely, the origin design parameters $\phi_{lens}$ (\eg, spherical interfaces, glass and air spacings, and the refractive index and Abbe number of the material) of the two applied MOS in $Zemax^\circledR$ software are directly fed into the simulation model, where the ISP is ignored, the patch size of the patch-wise convolution process is set to $16$, and the sensor pixel size is adjusted to $11.43 {\mu}m$ to match the reshaped image size. 
\add{The detailed descriptions of the optical simulation model can be found in the appendix.}
We select $2369$ images with a uniform resolution of $1024{\times}2048$ from Flickr2K~\cite{timofte2017ntire} as the ground truth for generating \textit{Syn}. Considering that the source domain data will not be applied for evaluation, \textit{Syn} contains only the training set.

\begin{figure}[!h]
  \centering
  \includegraphics[width=0.75\linewidth]{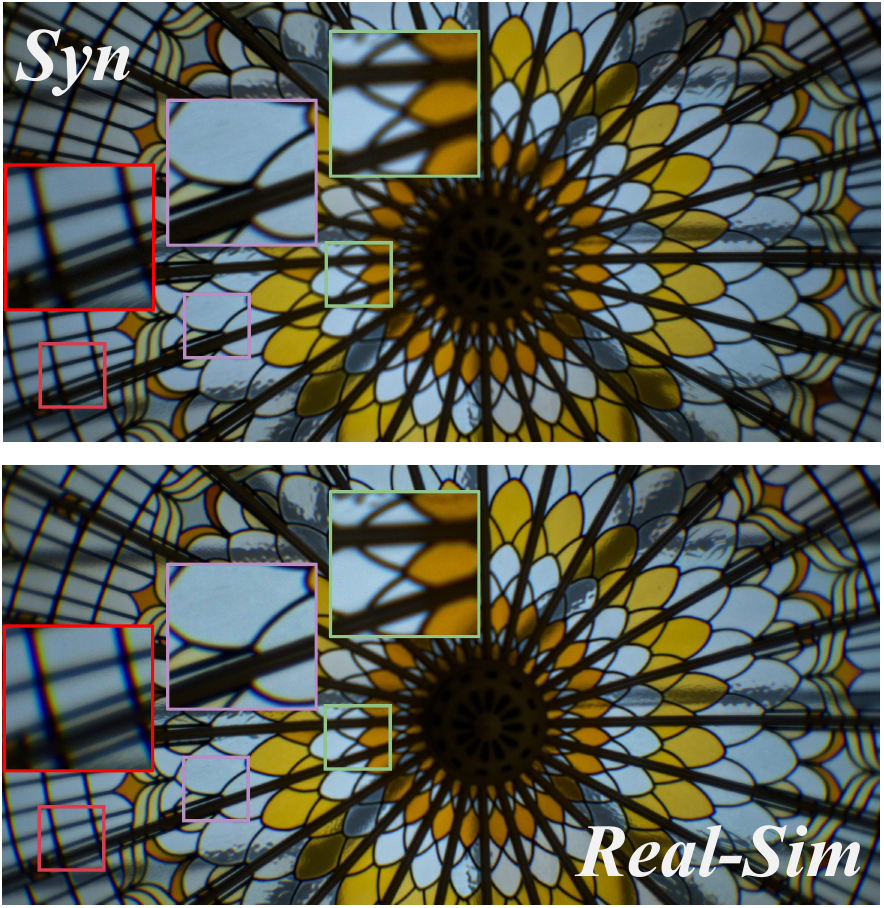}
  \caption{\textbf{Illustration of the simulated synthetic-to-real gap in \textit{Real-Sim}}. The simulated domain gap makes the degradation distribution of \textit{Real-Sim} deviate from that of \textit{Syn}.}
  \label{fig:gapsim}
\end{figure}

\PAR{Generation of \textit{Real-Sim}.}
Intuitively, investigating the DACAC task with snapped real-world aberrated images reveals a direct solution.
However, in this case, the evaluation protocol is limited by the no-referenced metrics (\eg~NIQE~\cite{mittal2012making}) and qualitative results, where the former can hardly assess the quality of the restored images, and the latter can only provide subjective assessment without objective numerical analysis.
In other words, it is challenging to design and study the UDA frameworks based on the unpaired real-world dataset. 
Moreover, preparing the paired real-world data remains a tough project in the field of CAC, which prevents the establishment of a convincing evaluation protocol with referenced metrics such as LPIPS~\cite{zhang2018unreasonable}, FID~\cite{heusel2017gans}, PSNR, and SSIM~\cite{wang2004image}.

Consequently, in Realab, we propose to simulate the synthetic-to-real gap for producing simulated ``real-world'' aberrated images paired with ground-truth images, \ie, \textit{Real-Sim}.
As shown in Figure~\ref{fig:gapsim}, the degradation distribution of \textit{Real-Sim} differs from that of \textit{Syn}, which indicates the existing domain gap between the two sets.
Despite that the target images are generated by the simulation model, the ground truth is only available during evaluation, which will not be applied for training.
In this way, we simulate the real-world situation, where the snapped aberrated images suffer deviated degradation distribution from the synthetic ones, whose ground truth is unavailable.
To be specific, we simulate the synthetic-to-real gap from the following aspects:
\begin{enumerate}
    \item We randomly adjust the origin design parameters $\phi_{lens}$ with a range of $\pm5\%$ to simulate \textit{the manufacture errors} that cause the aberration deviation (the random range is an empirical value, which can not only deliver an appropriate level of aberration deviation but also ensure that the optical system can be ray traced.).
    \item Compared to the larger patch size ($16$) in \textit{Syn}, the patch size is set to $8$ in \textit{Real-Sim}, which means \textit{a more accurate degradation process} for simulating the situation that no patch-wise approximation is performed in the actual imaging process.
    \item We consider \textit{the simulation of ISP} in generating \textit{Real-Sim}. The ground truth images are converted to raw images by the invert ISP~\cite{Chenshiqi} with the parameters of the applied sensor, which are randomly adjusted with a range of $\pm2\%$ (a larger range will lead to too many color shifts) in the following ISP pipeline. This process simulates the domain gap caused by the different ISP pipelines of the image sensors utilized to snap the ground-truth and aberrated images.
    \item The focal distance is adjusted to simulate \textit{the errors during the assembly of lenses}. In practical use, it is challenging to ensure that the focal distance of a lens mounted on a camera remains consistent with its design value, resulting in additional aberration deviation. Therefore, we adjust the original focal distance in the optical model by comparing the simulated results with the snapped images (using calibration boards), to produce the aberrated images under a realistic focal distance. 
    \item We select ground-truth images from another dataset DIV2K~\cite{timofte2017ntire} ($1024{\times}2048$), to ensure the \textit{differences in content} between \textit{Syn} and \textit{Real-Sim}. The image number of \textit{Real-Sim} is also reduced to $782$, which simulates the situation that the real-snapped images are often fewer than synthetic ones.
\end{enumerate}
Thanks to these efforts, \textit{Real-Sim} can be leveraged to benchmark CAC models under common synthetic-to-real gaps. 
A test set of $98$ images is set up to facilitate the qualitative evaluation under referenced metrics.

\PAR{Collection of \textit{Real-Snap}.}
To provide sufficient information about the target domain for UDA training and set up a reliable benchmark, the \textit{Real-Snap} covers rich and varied scenes of indoor, natural, campus, urban, and scenic spots. 
We snap $609$ and $561$ real-world aberrated images with MOS-S1 and MOS-S2 respectively, which are divided into $554/55$ and $508/53$ for the training/test set with a rough ratio of $10/1$.
The resolution of the images is also reshaped and cropped to $1024{\times}2048$.
For lack of ground-truth images, we evaluate the performance on the \textit{Real-Snap} with the widely-used no-reference metric NIQE~\cite{mittal2012making} and visual results.

\subsection{Implementation Details}
\label{subsec:details}
All the training processes of our work are implemented on two NVIDIA GeForce RTX 3090 GPUs through Adam optimizer of $\beta_{1} {=} 0.9, \beta_{2} {=} 0.99$, with a batch size of $8$.


\begin{figure}[!h]
  \begin{center}
  \includegraphics[width=0.65\linewidth]{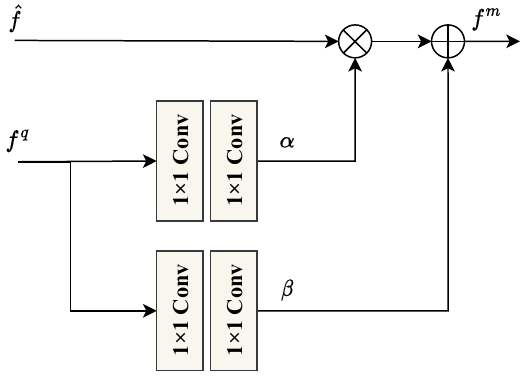}
  \end{center}
  \caption{\textbf{Illustration of the affine-based fusion module}.}
  \label{fig:affine}
\end{figure}

\PAR{\add{Network Architecture.}}
\add{For the encoder $E_{vq}, E_{cac}$ and decoder $G_{vq}, G_{cac}$, we adopt the same ResBlock-based~\cite{he2016deep} architecture as~\cite{chen2022real}.
As a common practice, the U-Net discriminator with spectral normalization~\cite{wang2021real,wu2023ridcp} is applied for ${D}_{vq}$, ${D}_{cac}$, and ${D}_{fa}$. 
We note that the proposed QDMR is a framework approach with no special requirement for the network structure.
For the bottleneck module $B_{cac}$, any powerful backbone for low-level vision is applicable.
Thereby, in the following experiments, the residual Swin transformer layers (Swin)~\cite{liang2021swinir} and residual-in-residual dense block (RRDB)~\cite{wang2018esrgan} are selected for examples.
The number of both Swin and RRDB blocks is set to $4$ the same as~\cite{chen2022real}. 
The codebook size of DMC is set as $K=1024$ and the dimension of the feature is set to $n=512$ following~\cite{wu2023ridcp,chen2022real}. 
Before and after the feature quantization, we apply a $1\times1$ convolution layer to process the restoration feature and align the deep feature dimension ($256$) with the codebook dimension ($512$). 
Meanwhile, the implementation of the affine-based fusion module is illustrated in Figure~\ref{fig:affine}.}

\PAR{\add{Training Details.}}
\add{Different from $\mathcal{L}_{s2tT}$ and $\mathcal{L}_{fa}$, following~\cite{wang2021real,chen2022real,wu2023ridcp}, we adopt the hinge loss for $\mathcal{L}_{adv}$ in $\mathcal{L}_{cac}$ to generate realistic clear images. 
A pre-trained VGG-16 network is utilized for calculating the $\mathcal{L}_{per}$.
The loss weights are set as $\lambda_{per}^{vq}{=}1, \lambda_{adv}^{vq}{=}0.1$, $\lambda_{per}^{cac}{=}1$, $\lambda_{adv}^{cac}{=}0.01$, $\lambda_{s}{=}1$, $\lambda_{s2t}{=}1$, $\lambda_{t}{=}0.1$, $\lambda_{fa}{=}0.01$, empirically for a stable training process.  
For data augmentation, random crop, flip, and rotation are applied, where the crop size is $256{\times}256$.
During the training stage of VQGAN, the model is trained for $200K$ iterations with a fixed learning rate of $1e{-}4$, on both synthetic and real-world aberrated images $\{y_{S}, y_{R}\}$ by optimizing $\mathcal{L}_{vq}$ and $\mathcal{L}_{s2tT}$. 
The parameters of the image reconstruct network and the DMC are updated during this stage.
Then, the parameters of DMC are utilized during the CAC stage for feature quantization and fusion, which are frozen to store the learned degradation priors, and the whole VQGAN is applied to produce pseudo paired target data $\mathcal{D}_{S2T}$. 
We train the CAC model for $200K$ iterations with an initial learning rate of $1e{-}4$, which is halved at $100k$, $150k$, $175k$, and $190k$.
For the training of QDMR-Base, we optimize $L_{cac}$ on the synthetic data pairs $\mathcal{D}_{S}$.
For the training of QDMR-UDA, the training objective $L_{uda}$ is optimized on both synthetic data pairs $\mathcal{D}_{S}$, pseudo paired target data $\mathcal{D}_{S2T}$, and real-world data $\mathcal{D}_{T}$. }

\begin{table*}[h!]
    \begin{center}
        \caption{\textbf{Quantitative Comparison with Source-only CAC Models.} The ``Source-only'' means that the models are trained only on source data \textit{Syn}. ``RRDB'' and ``Swin'' represent the applied backbones for $B_{cac}$. The \YKL{best} and \JQ{second} results are highlighted (except for the oracle performance). }
        \label{tab:restore}
        \input{compare_restore}
    \end{center}
\end{table*}

\PAR{Implementations of Competing Methods.}
In order to make the methods in other low-level tasks applicable to the DACAC task, we modify the implementations for some of them.
Following~\cite{jiang2023minimalist}, the pixel-unshuffle operation is applied for the input image in all SR methods~\cite{wang2018esrgan,liang2021swinir,zhou2023srformer}, to reduce the size of the processed image to avoid excessive computational overhead. 
Then, for PiRN-Refine~\cite{jaiswal2023physics}, we only follow its idea of utilizing the diffusion model to refine the restoration results of the trained PSNR-oriented model.
Considering that the code of PiRN-Refine is not publicly available yet, the trained SwinIR is used as a preliminary restoration model, where the SR3~\cite{saharia2022image} is then trained to refine its results.
In terms of all competing UDA methods~\cite{wei2021unsupervised,guo2020closed,wang2021unsupervised}, we only care about the performance of their training framework, so all networks are replaced with our baseline structure, \ie, QDMR-Base model without codebook quantization and fusion module.
Meanwhile, in DASR~\cite{wei2021unsupervised}, the bicubic downsample process in DSN training is replaced by the image simulation in generating \textit{Syn} to fit the DACAC task.

\subsection{Results on Synthetic Benchmark: \textit{Syn$\to$Real-Sim}}
\label{subsec:syn}
\begin{figure*}[!t]
  \centering
  \includegraphics[width=0.8\linewidth]{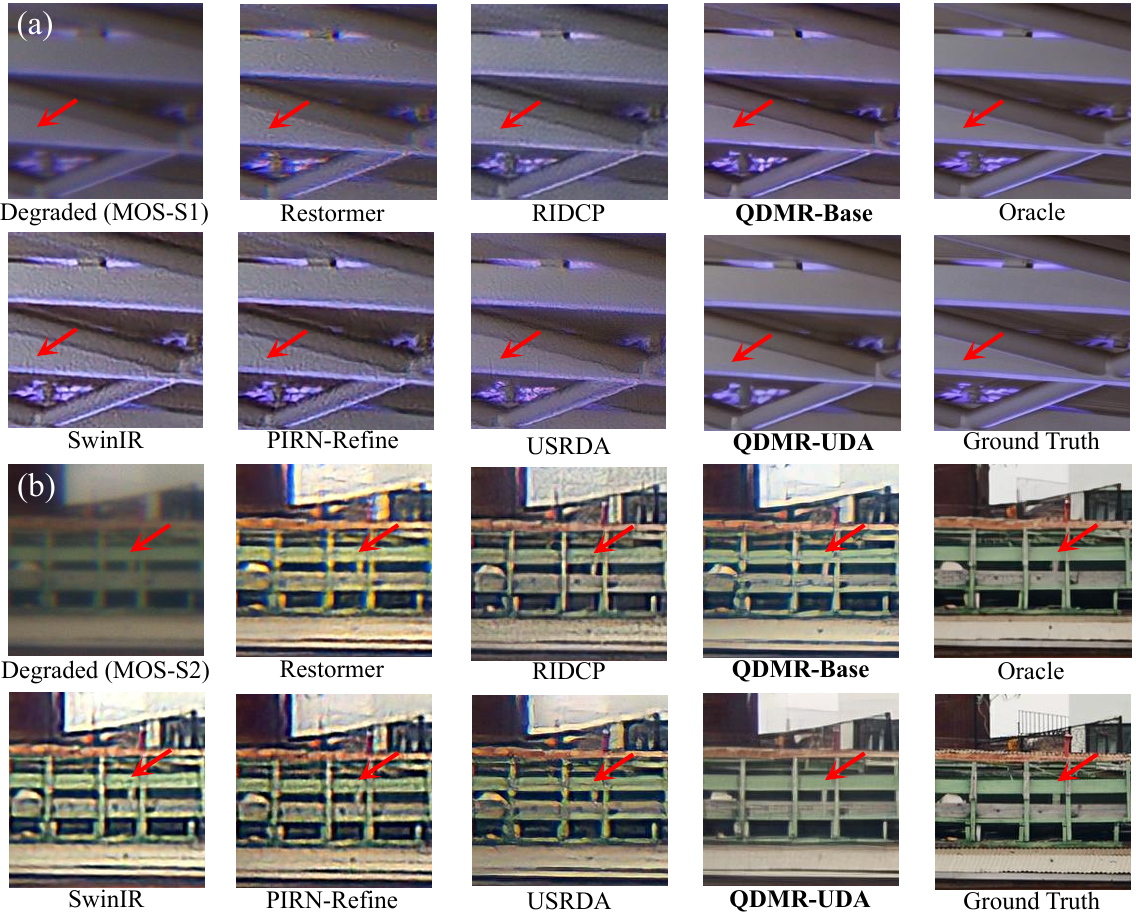}
  \caption{\add{\textbf{Visual results on \textit{Real-Sim}.}} (a) and (b) are results for MOS-S1 and MOS-S2. While the restored images of most competing methods suffer from severe artifacts, the proposed QDMR-based methods can deliver realistic CAC results comparable to the oracle performance and the ground truth. Please zoom in for the best view.}
  \label{fig:syn-results}
\end{figure*}

We first report both quantitative and qualitative results on the proposed synthetic benchmark \textit{Syn$\to$Real-Sim} to conduct a preliminary evaluation of competing methods.
Considering that the DACAC task aims to generate images of realistic textures and reduce the domain-gap-induced artifacts, in Table~\ref{tab:restore} and Table~\ref{tab:uda}, we mainly focus on the perceptual-based metric LPIPS and FID, while employing fidelity-based metrics PSNR and SSIM for auxiliary references.
The visual results of the most competitive methods in the tables are shown in Figure~\ref{fig:syn-results}. 


\PAR{Comparison with Source-only CAC Models.}
The QDMR-Base model is compared to other potential image restoration models which can be applied in CAC, \ie, SR models~\cite{wang2018esrgan,liang2021swinir,zhou2023srformer}, Deblur models~\cite{chen2022simple,zamir2022restormer,wang2022uformer}, and generative models of GAN-based methods~\cite{wang2018esrgan,liang2022details}, diffusion-based method~\cite{jaiswal2023physics} and VQ-based methods~\cite{chen2022real,wu2023ridcp}. 
To maintain a fair comparison, all the compared models are retrained on the source domain data $\mathcal{D}_{S}$ under two aberration behaviors, with their default training strategies but keeping the same iterations and batch size with our model, and tested on the corresponding test set of the target domain data $\mathcal{D}_{T}$. 
The oracle performance where the baseline CAC model is supervised directly on paired target images is also provided as the upper limit of the DACAC task. 
As illustrated in Table~\ref{tab:restore}, suffering from the synthetic-to-real domain gap, all competing models reveal unsatisfactory performance on two MOS samples, where the LPIPS deteriorates by $0.181{\sim}0.248$ (about $163\%{\sim}223\%$) for MOS-S1 and $0.217{\sim}0.304$ (about $129\%{\sim}181\%$) for MOS-S2 compared to the oracle performance. 
Among them, generative models, which are commonly applied in real-world image restoration, generally yield relatively better results, but the improvements are rather limited due to the lack of information about the target domain. 
The proposed QDMR-Base model achieves far superior LPIPS under all aberration behaviors, which also delivers impressive visual results with realistic textures and fewer artifacts as shown in Figure~\ref{fig:syn-results}. 
The results provide sufficient evidence that the guidance of QDMR is beneficial for enhancing the adaptation of the CAC model with different architectures to the target domain. 

\begin{table*}[h!]
    \begin{center}
        \caption{\textbf{Quantitative Comparison with UDA Methods.} Read as Table~\ref{tab:restore}.}
        \label{tab:uda}
        \input{compare_UDA}
    \end{center}
\end{table*}

\PAR{Comparison with Existing UDA Methods.}
Then, we explore the DACAC task by incorporating the target domain data $\mathcal{D}_{T}$ for UDA training.
Considering that no efforts have been made in this setting, we compare our QDMR-UDA framework against existing UDA methods in other low-level vision tasks. Since most of them are task-specific or publicly unavailable, we compare our framework with DASR~\cite{wei2021unsupervised}, DRN-Adapt~\cite{guo2020closed}, and USRDA~\cite{wang2021unsupervised}, which are powerful and universal UDA frameworks that can be adopted in DACAC. 
These frameworks are trained with the same setting as QDMR-UDA, where the network architecture is also kept the same as our baseline CAC model for a fair comparison.
Despite being equipped with domain transformation and adversarial feature alignment strategies, the competing UDA frameworks in Table~\ref{tab:uda} fail to transfer the relevant knowledge of the target domain to the CAC model, only bringing limited improvements over the source-only models in Table~\ref{tab:restore}. 
In contrast, the proposed QDMR effectively stores the optical degradation priors in the DMC and provides valuable guidance for the CAC model.
Even the QDMR-Base model delivers superior results over the compared UDA methods. 
Moreover, our QDMR-UDA framework further unlocks the potential of QDMR, achieving exceptional performance in all metrics.
Specifically, QDMR-UDA improves QDMR-Base at most by $0.083$ in LPIPS, $24.45$ in FID, $4.94dB$ in PSNR, and $0.102$ in SSIM.
The visual results of QDMR-UDA in Figure~\ref{fig:syn-results} are also more visually pleasant with almost no artifacts, which are close to the oracle performance and the ground truth.

\subsection{Results on Real-world Benchmark: \textit{Syn$\to$Real-Snap}}
\label{subsec:real}
\begin{figure*}[!t]
  \centering
  \includegraphics[width=1.0\linewidth]{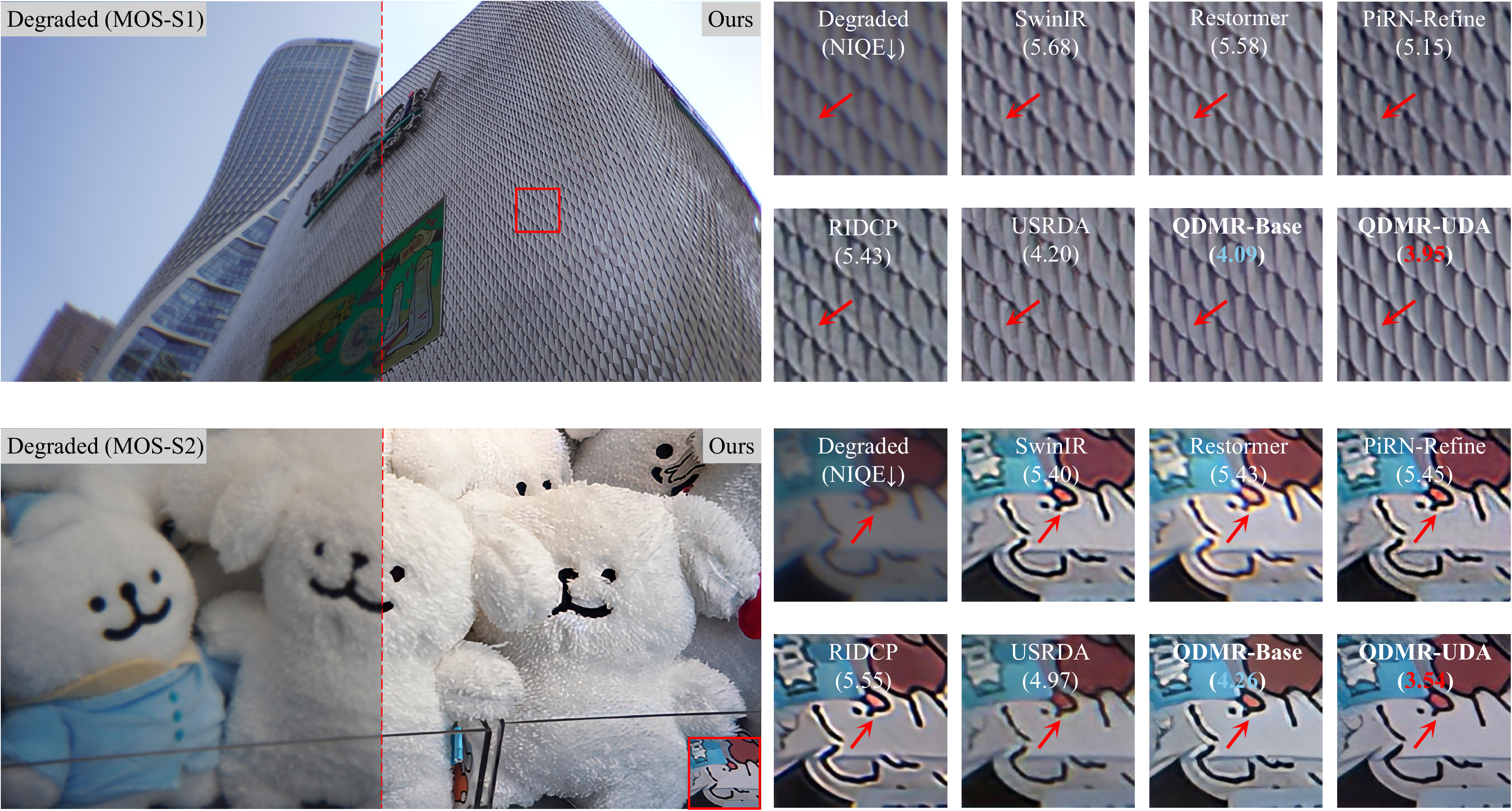}
  \caption{\add{\textbf{Visual results on \textit{Real-Snap}.}} \add{Similar to the performance on the synthetic benchmark \textit{Real-Sim},} benefiting from the learned QDMR and applied UDA strategies, QDMR-UDA restores more realistic textures and structures while suppressing the artifacts, achieving the best NIQE. Please zoom in for the best view.}
  \label{fig:real-results}
\end{figure*}

The best-performing methods on the synthetic benchmark, \ie, SwinIR~\cite{liang2021swinir} in SR, Restormer~\cite{zamir2022restormer} in Deblur, PIRN-Refine~\cite{jaiswal2023physics} and RIDCP~\cite{wu2023ridcp} in Generative models, USRDA~\cite{wang2021unsupervised} in UDA methods, and our proposed QDMR-Base-Swin and QDMR-UDA-Swin, are selected to evaluate their performance on the real-world benchmark.
The source-only models are directly tested on the test set of \textit{Real-Snap} considering that they can not be trained on the unpaired target images. 
For USRDA, QDMR-Base, and QDMR-UDA, we replace the synthetic target images with real-world ones for training new models. 
The qualitative results on the real-world benchmark \textit{Syn$\to$Real-Snap} of both aberration behaviors are shown in Figure~\ref{fig:real-results}, along with NIQE ($\downarrow$) of each method. 
\add{Consistent with the results on the synthetic benchmark, UDA training reveals an excellent solution to the synthetic-to-real gap}, and QDMR-based methods outperform the competing methods by a large margin, where QDMR-UDA and QDMR-Base improve NIQE up to $36\%$ and $28\%$ respectively.  
\add{To be specific, similar to those on \textit{Real-Sim}, the output images of source-only methods exhibit artifacts and color shifts in flat areas, while suffering distortions and color fringing at the edges. In contrast, the UDA methods, especially our QDMR-based methods, showcase smoother results in flat regions, with natural and uniform colors, while also presenting sharp and undiffused edges.}
In a word, QDMR-UDA and QDMR-Base tend to produce CAC results with clearer details, higher contrast, fewer artifacts, and fewer uncorrected aberrations.
\add{Additionally, we find that the forms of artifacts observed in the real-world benchmark differ from those in the synthetic one. The former often manifests as distortions in shapes located at the edges, while the latter resembles noise-like textures. However, these issues do not appear in the results of the proposed QDMR-UDA, indicating its superior ability to address complex and unknown synthetic-to-real gaps in CAC.}

The experimental results on both benchmarks demonstrate that the proposed QDMR can effectively mitigate the domain gap issue in real-world CAC. 
More visual samples can be found in the appendix. 

\subsection{Ablation Study}
\label{subsec:ablation}
We conduct ablation studies to investigate how to learn useful QDMR, how to adapt the CAC model to the target domain with its guidance, and the impacts of the target data scale on DACAC results. 
All the ablations are implemented on \textit{Syn$\to$Real-Sim} setting of MOS-S1 with Swin for $B_{cac}$, where LPIPS/PSNR is provided as the reference metrics. 
\begin{table}[t!]
    \begin{center}
        \caption{\textbf{Ablations on Components of QDMR-UDA.} ``\textit{w/o}'': restoring image directly from $f^q$. ``DCN'': deformable convolution network~\cite{zhu2019deformable}. ``MCA'': multi-modal cross attention~\cite{sun2022event}. ``Affine'': applied affine-based feature modulation.}
        \label{tab:ab_uda}
        \input{ablation_uda}
    \end{center}
\end{table}

\PAR{Components of QDMR-UDA.}
In Table~\ref{tab:ab_uda}, based on the baseline model depicted in Section~\ref{subsec:qdmrbase}, we gradually add components of QDMR-UDA to study their respective contributions.
Besides, different methods for fusing the quantized priors $f^q$ with the restoration feature are also explored. 
As shown in setting $1{\sim}6$, the learned QDMR provides effective guidance for the CAC model with simple fusion strategies, where the model achieves superior performance (improvements of $0.022~(8\%)$ in LPIPS) equipped with the Affine module. 
The adversarial loss and feature alignment prove to be powerful strategies (setting $7{\sim}8$) to further enhance the domain generalization ability of the QDMR-guided model.
Remarkably, the s2tT makes significant contributions to the final results by producing the data flow of $\mathcal{D}_{S2T}$, bringing improvements of $0.043~(19\%)$ in LPIPS and $1.4dB~(15\%)$ in PSNR (setting $8{\sim}9$).

\begin{table}[t!]
    \begin{center}
    \caption{\textbf{Ablations on Pretrain Phase of VQGAN.} ``baseline'': the vanilla training objective in Eq.~\ref{eq:vqganloss_s}. ``FA'': adding $\mathcal{L}_{fa}$ to Eq.~\ref{eq:vqganloss_s}. ``s2tT'': replacing $\mathcal{L}_{adv}$ in Eq.~\ref{eq:vqganloss_s} with $\mathcal{L}_{s2tT}$. ``\textit{all}'': applying both $\mathcal{L}_{fa}$ and $\mathcal{L}_{s2tT}$.}
    \label{tab:ab_pretrain}
    \input{ablation_pretrain}
     \end{center}
\end{table}

\PAR{Pretrain Phase of VQGAN.}
The effective QDMR and s2tT of the VQGAN is the key to achieving the impressive results. We further investigate the data usage and constraints for training a powerful VQGAN in Table~\ref{tab:ab_pretrain}.
It is observed that learning degradation-aware priors, shows comparable performance with learning HQPs (row 2 and row 3), while the representation for the target domain contributes to a better result (row 3 and row 4).
However, the mixing of data from both domains is essential to a better understanding of the domain gap, bringing generally high performance (rows 5-8).
More importantly, it enables the crucial s2tT ability of VQGAN for UDA training.
In terms of training objectives, applying the proposed $\mathcal{L}_{s2tT}$ alone (row 7) is the optimal solution, which strikes a fine balance between QDMR learning and s2t data generation. 

\begin{table}[t!]
    \caption{\textbf{Ablations on Combination of Data Flow.} The $\mathcal{D}_{s2t}$ for baseline is also generated by the VQGAN of QDMR. ``$\mathcal{D}_{S}$'': $\lambda_{s2t} {=} 0$,  $\lambda_{fa} {=} 0$. ``$\mathcal{D}_{S}\&\mathcal{D}_{T}$'': $\lambda_{s2t} {=} 0$. ``$\mathcal{D}_{s2t}$'': $\lambda_{s} {=} 0$, $\lambda_{fa} {=} 0$.}
    \label{tab:ab_s2t}
    \input{ablation_s2t}
\end{table}

\PAR{Combination of Data Flow.} 
Table~\ref{tab:ab_s2t} reports the results of the baseline model and QDMR-based model under different combinations of data flow, which are implemented by adjusting $\lambda_{s}$, $\lambda_{s2t}$ and $\lambda_{fa}$ in Eq.~\ref{eq:total}. 
The generated pseudo paired set $\mathcal{D}_{s2t}$ provides convincing supervision on the target domain, delivering superiority over other data flows (rows 1-3). 
Moreover, when all the data flows are incorporated into training by $\mathcal{L}_{uda}$, the models yield excellent results (row 4). 
Notably, in this case, even the baseline model can achieve a remarkable performance of $0.190/24.54dB$ in LPIPS/PSNR, while the learned QDMR can further improve the baseline by $0.008$ in LPIPS and $0.92dB$ in PSNR.

\begin{figure}[!t]
  \begin{center}
  \includegraphics[width=0.8\linewidth]{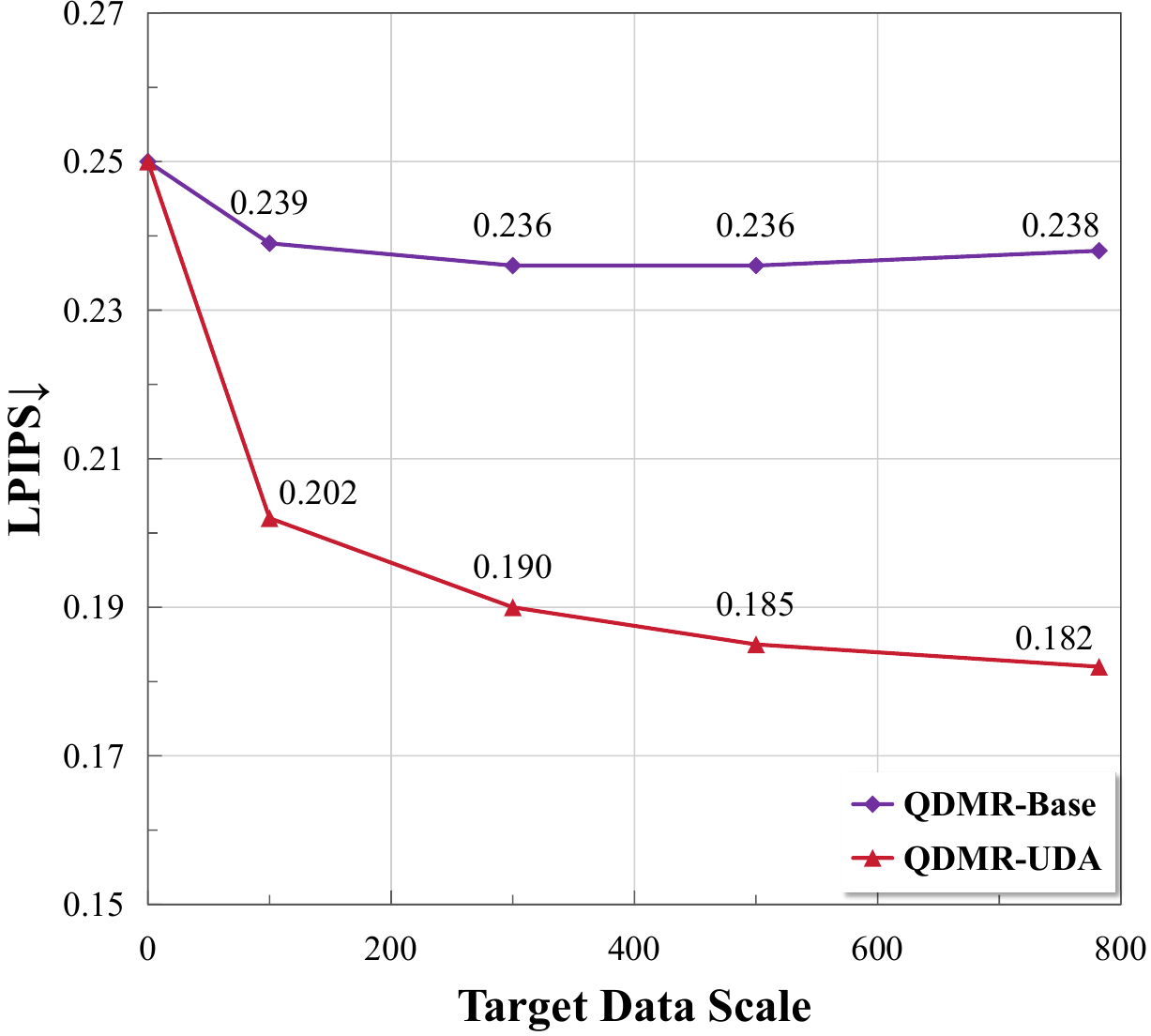}
  \end{center}
  \caption{\textbf{Impacts of the target data scale}. We randomly select $100$, $300$, and $500$ samples from the entire target training set to retrain our methods.}
  \label{fig:datascale}
\end{figure}

\PAR{Target Data Scale.}
With the recognition of the considerable time and labor investment in snapping real-world aberrated images, it is essential to explore the target data scale required for effective UDA training. 
Figure~\ref{fig:datascale} reports the performance of QDMR-Base and QDMR-UDA trained with different numbers of target images.
It can be observed that only a few target images are needed for learning effective QDMR, where the target data scale even reveals negligible impacts on the results of QDMR-Base, providing evidence for the few-shot capabilities of QDMR.
Thanks to the UDA training strategies, a larger target data scale seems beneficial to improving QDMR-UDA.
However, the performance of QDMR-UDA with $500$ training samples is comparable to that of the whole training set. 
In this way, to strike a balance between the efforts devoted to dataset collection and the effectiveness of UDA training, the training data scale for \textit{Real-Snap} of about $500$ is reasonable.

%% file: compare_restore.tex
\resizebox{0.8\textwidth}{!}
{
\renewcommand{\arraystretch}{1.2}
\setlength{\tabcolsep}{1mm}{
\begin{tabular}{cc|cccc|cccc}
\hline
\multicolumn{2}{c|}{\multirow{2}{*}{\begin{tabular}[c]{@{}c@{}} \textbf{\textit{Syn$\to$Real-Sim}}\\ Method\end{tabular}}} & \multicolumn{4}{c|}{\textbf{MOS-S1}} & \multicolumn{4}{c}{\textbf{MOS-S2}} \\ \cline{3-10} 
\multicolumn{2}{c|}{} & LPIPS$\downarrow$ & FID$\downarrow$ & PSNR$\uparrow$ & SSIM$\uparrow$ & LPIPS$\downarrow$ &FID$\downarrow$ &  PSNR$\uparrow$ & SSIM$\uparrow$   \\ \hline \hline
\multirow{3}{*}{SR} 
 & RRDBNet~\cite{wang2018esrgan} &0.350 &36.28 &21.63  &0.725  &0.472 &82.48  &17.35  &0.581  \\
 & SwinIR~\cite{liang2021swinir} &0.334 &\YKL{31.63}  &22.24  &0.736  &0.426 &74.40  &17.12  &0.631  \\
 & SRFormer~\cite{zhou2023srformer} &0.341 &35.75  &21.74  &0.709 &0.422 &74.38  &17.17  &\JQ{0.640}  \\ \hline
\multirow{3}{*}{Deblur} & NAFNet~\cite{chen2022simple} &0.353  &41.97  &\JQ{23.15}  &0.718   &0.419 &71.86   &17.46  &\YKL{0.649}  \\
 & Restormer~\cite{zamir2022restormer} &0.311 &35.73  &23.12  &\YKL{0.749}  &0.431 &\YKL{68.25}  &17.07  &0.627  \\
 & UFormer~\cite{wang2022uformer} &0.326 &33.74  &22.13  &0.713   &0.432 &75.22  &17.51  &0.617  \\ \hline
\multirow{5}{*}{\makecell[c]{Generative\\Model}} & ESRGAN~\cite{wang2018esrgan} &0.316 &44.14  &21.79  &0.627  &0.436 &96.13  &17.42  &0.432  \\
 & SwinIR \textit{w} LDL~\cite{liang2022details} &0.292 &42.19  &21.38  &0.659    &0.401 &96.26  &16.14  &0.495  \\
 & PiRN-Refine~\cite{jaiswal2023physics} &0.297 &35.25  &22.71  &0.678   &0.385 &74.31  &\YKL{19.20}  &0.596  \\
 & FeMaSR~\cite{chen2022real} &0.310 &44.80  &22.65  &0.662  &0.410  &90.19  &17.18  &0.491  \\
 & RIDCP~\cite{wu2023ridcp} &0.301 &38.53  &22.73  &0.720  &0.392  &\JQ{70.18} 
 &\JQ{19.12}  &0.590 \\ \hline
\multirow{2}{*}{\textbf{\makecell[c]{QDMR-Base\\(Ours)}}} 
 & RRDB &\JQ{0.261} &39.29  &\YKL{23.43}  &\JQ{0.737}  &\JQ{0.330}  &71.45  &18.65  &0.608  \\
 & Swin &\YKL{0.238} &\JQ{32.75}  &22.75  &0.699   &\YKL{0.315} &73.74  &17.76  &0.602  \\ \hline
\multirow{2}{*}{{\makecell[c]{Oracle}}} 
 &  RRDB &0.165 &18.20   &26.63  &0.842  &0.211 &25.51   &24.61  &0.816  \\
  & Swin &0.111 &11.99   &30.21  &0.876  &0.168 &18.33   &28.02  &0.829   \\ \hline
\end{tabular}

}
}

%% file: compare_UDA.tex
\resizebox{0.8\textwidth}{!}
{
\renewcommand{\arraystretch}{1.2}
\setlength{\tabcolsep}{1mm}{
\begin{tabular}{cc|cccc|cccc}
\hline
\multicolumn{2}{c|}{ \textbf{\textit{Syn$\to$Real-Sim}}} & \multicolumn{4}{c|}{\textbf{MOS-S1}} & \multicolumn{4}{c}{\textbf{MOS-S2}} \\ \cline{3-10} 
\multicolumn{2}{c|}{Method} & LPIPS$\downarrow$ & FID$\downarrow$ & PSNR$\uparrow$ & SSIM$\uparrow$  & LPIPS$\downarrow$ & FID$\downarrow$ & PSNR$\uparrow$ & SSIM$\uparrow$   \\ \hline\hline
  \multirow{2}{*}{DASR~\cite{wei2021unsupervised}} & RRDB &0.281 &38.26 &22.57  &0.706  &0.381 &68.60 &19.70    &0.637  \\
   & Swin &0.277 &39.35  &23.05  &0.714  &0.371 &62.39 &19.58  &0.615    \\
  \multirow{2}{*}{DRN-Adapt~\cite{guo2020closed}} & RRDB &0.311  &33.23  &23.56  &0.736    &0.396 &73.30   &20.00  &0.632  \\
   & Swin &0.305 &\JQ{30.34}  &\JQ{24.33}  &0.733  &0.377 &68.45 &19.47  &0.625    \\
  \multirow{2}{*}{USRDA~\cite{wang2021unsupervised}} & RRDB &0.281 &38.14  &23.36  &0.723  &0.355 &53.52  &\JQ{21.51}  &0.679  \\
   & Swin &0.279 &31.78 &23.78  &0.700  &0.351 &66.18  &18.74  &0.600   \\ \hline
\multirow{2}{*}{\textbf{\makecell[c]{QDMR-Base\\(Ours)}}} 
 & RRDB &{0.261} &39.29  &23.43  &0.737  &{0.330} &71.45  &18.65  &0.608  \\
 & Swin &{0.238} &32.75  &22.75  &0.699   &{0.315} &73.74  &17.76  &0.602  \\ \hline
\multirow{2}{*}{\textbf{\makecell[c]{QDMR-UDA\\(Ours)}}} 
 &  RRDB &\JQ{0.225} &35.42  &23.79  &\JQ{0.738}  &\JQ{0.252} &\JQ{47.42}   &20.93  &\JQ{0.693}   \\
  & Swin &\YKL{0.182} &\YKL{26.54}  &\YKL{25.46}  &\YKL{0.767}  &\YKL{0.232} &\YKL{44.15}  &\YKL{22.70}  &\YKL{0.704}  \\ \hline
\multirow{2}{*}{{\makecell[c]{Oracle}}} 
 &  RRDB &0.165 &18.20  &26.63  &0.842  &0.211 &25.51  &24.61  &0.816  \\
  & Swin &0.111 &11.99  &30.21  &0.876  &0.168 &18.33  &28.02  &0.829   \\ \hline

\end{tabular}

}
}

%% file: ablation_uda.tex
\resizebox{0.5\textwidth}{!}
{
\renewcommand{\arraystretch}{1.2}
\setlength{\tabcolsep}{1mm}{
\begin{tabular}{cccccc|cc}
\hline
         & \textbf{QDMR} & \textbf{Fusion}       & \bm{$\mathcal{L}_{adv}$} & \bm{$\mathcal{L}_{fa}$}   & \bm{${\mathcal{D}_{S2T}}$}  & LPIPS$\downarrow$ & PSNR$\uparrow$  \\ \hline\hline
baseline & \ding{55} & \textit{n.a.}         & \ding{55}    & \ding{55}  & \ding{55}   &0.286 &22.97      \\
1        & \checkmark  & \textit{w/o} & \ding{55}    & \ding{55}     & \ding{55}           &0.270 &22.59         \\
2        & \checkmark  & Add          & \ding{55}    & \ding{55}     & \ding{55}           &0.310  &22.95          \\
3        & \checkmark  & Concat.      & \ding{55}    & \ding{55}     & \ding{55}           &0.283  &23.93           \\
4       & \checkmark   &DCN      & \ding{55}    & \ding{55}     & \ding{55}           &0.305  &22.93           \\
5        & \checkmark  &MCA      & \ding{55}    & \ding{55}     & \ding{55}           &0.311  &22.85           \\
6        & \checkmark  & Affine      & \ding{55}    & \ding{55}     & \ding{55}            &0.264  &22.87         \\
7        & \checkmark  & Affine      & \checkmark     & \ding{55}     &\ding{55}           &0.238  &22.75         \\
8        & \checkmark  & Affine      & \checkmark & \checkmark     & \ding{55}               &0.225 &24.06        \\
9        & \checkmark  & Affine      & \checkmark & \checkmark     & \checkmark             &\textbf{0.182}  &\textbf{25.46}         \\
\hline
\end{tabular}

}
}

%% file: ablation_pretrain.tex
\resizebox{0.35\textwidth}{!}
{
\renewcommand{\arraystretch}{1.2}
\setlength{\tabcolsep}{1mm}{

\begin{tabular}{cc|cc}
\hline
\multicolumn{2}{c|}{\textbf{Pretrain Phase}} & \textbf{Base}  & \textbf{UDA}       \\ \hline \hline
\multicolumn{2}{c|}{\textit{w/o}}      &0.267/23.02           &\textit{n.a.}  \\
\multicolumn{2}{c|}{HQP~\cite{chen2022real}}                &0.244/22.87           & \textit{n.a.} \\
\multicolumn{2}{c|}{Source-only}           &0.250/22.60          & \textit{n.a.}               \\
\multicolumn{2}{c|}{Target-only}           &0.242/22.44        & \textit{n.a.}               \\
\multirow{4}{*}{Mixing}  & baseline &0.242/22.68         &0.214/24.03    
                \\
                         & FA  &0.232/22.47        &0.207/23.20  
              \\
                         & s2tT &0.238/22.75        &\textbf{0.182}/\textbf{25.46}    
             \\
                         & \textit{all} &0.234/23.07 &0.210/24.72 
              \\
\hline
\end{tabular}

}
}

%% file: ablation_s2t.tex
\resizebox{0.35\textwidth}{!}
{
\renewcommand{\arraystretch}{1.2}
\setlength{\tabcolsep}{1mm}{
\begin{tabular}{c|cc}
\hline
\textbf{Data Flow} & \textbf{Baseline} & \textbf{QDMR} \\ \hline\hline
$\mathcal{D}_{S}$               & 0.267/23.02       & 0.238/22.75   \\
$\mathcal{D}_{S}\&\mathcal{D}_{T}$           & 0.224/24.05       & 0.225/24.06   \\
$\mathcal{D}_{S2T}$               & 0.199/23.77                 & 0.190/23.72   \\
\textit{all}       & 0.190/24.54                 & \textbf{0.182}/\textbf{25.46}   \\ \hline
\end{tabular}

}
}

%% file: conclusion.tex
In this paper, the real-world CAC is formulated as a UDA problem to mitigate the synthetic-to-real gap, where we deliver a comprehensive dataset Realab to facilitate research in this area. 
A novel QDMR method is proposed to characterize the domain gap information into a DMC by training a VQGAN. 
We design the QDMR-Base model where the CAC feature is modulated by the learned DMC and further introduce the QDMR-UDA framework to unlock the potential of QDMR in domain data transformation and feature alignment. 
The experimental results on both synthetic and real-world benchmarks demonstrate the effectiveness of QDMR in adapting the CAC model to the target domain with unpaired real-world aberrated images.
\add{In summary, developing UDA solutions integrating the unpaired real-world images in the training process contributes a lot to achieving comparable CAC results in real-world scenes as on synthetic data.}
\add{In this way, our work is expected to offer a unique perspective for the computational photography community in facilitating real-world applications of CAC methods.}

\PAR{Limitations and Future Work.}
However, some limitations require further exploration. 
Firstly, the few-shot capability of the UDA framework needs to be further developed.
The main drawback of the UDA solution lies in the requirements for the real-snapped images, which will consume a certain amount of time and effort.
How to learn the degradation distribution of the real-world target domain with fewer data (dozens or even several pieces) to achieve excellent CAC results will be an important research direction of the DACAC task.

Then, the scheme of learning degradation-aware priors does not perform well in dealing with some high-frequency details and the restoration of text symbols. We hope to introduce high-quality priors in future work and explore the integration of the two types of priors to achieve better CAC results.

Last but not least, in order to enlarge and simulate the domain gap to showcase its impacts on CAC, we did not calibrate and refine the parameters of the optical model in \textit{Syn} generation. 
\add{Additionally, the arbitrary image restoration solution was also not discussed in our work, where a universal model is developed based on training on a large scale of data under diverse optical degradation.}
However, in practical applications, \add{there is no conflict between building up a robust CAC model and applying the UDA framework.}
\add{The combination of a high-precision simulation pipeline, large CAC model with strong generalization ability}, and powerful UDA framework is expected to approach the upper limit of real-world CAC.

%% file: supp.tex
\section{More Details on Imaging Simulation}
\label{sec:data}
\begin{figure*}[!h]
  \centering
  \includegraphics[width=0.7\linewidth]{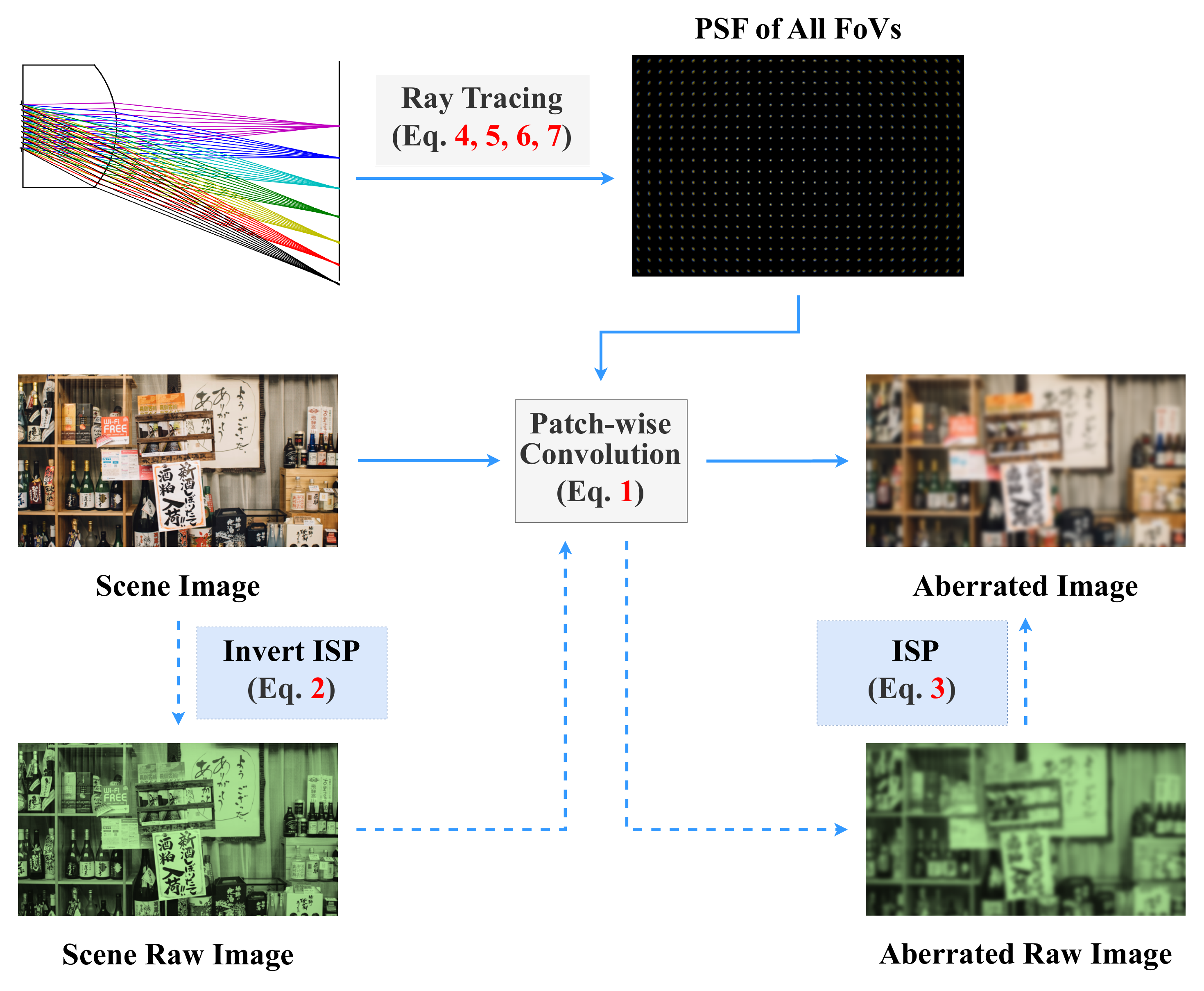}
  \caption{\textbf{The imaging simulation pipeline.} With the parameters of the applied lens and ISP, we can generate synthetic aberrated images through the ray-tracing-based simulation model.}
  \vskip-2ex
  \label{fig:gy_simu}
\end{figure*}
In our imaging simulation pipeline, the degradation caused by aberrations is characterized by the energy dispersion of the point spread function $PSF(x',y')$, where $(x',y')$ represents image plane coordinates. As shown in Figure~\ref{fig:gy_simu}, our simulation model applies the patch-wise convolution between the scene image $I_S$ and $PSF(x',y')$ for generating the aberrated image $I_A$:
\begin{equation}
\label{eq:gaoyao1}
    I_A(x',y') \approx PSF(x',y') \ast I_s(x',y').
\end{equation}
Besides, an optional Image Signal Processing (ISP) pipeline is introduced to help construct more realistic aberrated images~\cite{brooks2019unprocessing}. In this case, the scene image $I_S$ is replaced by the scene raw image $I'_S$, and the aberrated image $I_A$ is replaced by the aberrated raw image $I'_A$. Specifically, we first sequentially apply the invert gamma correction (GC), invert color correction matrix (CCM), and invert white balance (WB) to $I_S$ to obtain the scene raw image $I'_S$. The invert ISP pipeline can be formulated as
\begin{equation}
\label{eq:gaoyao2}
    I'_S = P^{-1}_{WB} \circ P^{-1}_{CCM} \circ P^{-1}_{GC}(I_S),
\end{equation}
where $\circ$ is the composition operator. $P_{WB}$, $P_{CCM}$, and $P_{GC}$ represent the procedures of WB, CCM, and GC, respectively. After conducting patch-wise convolution with the $PSF(x',y')$, we mosaic the degraded raw image $I'_A$ before adding shot and read noise to each channel. Moreover, we sequentially apply the demosaic algorithm, \ie, WB, CCM, and GC, to the R-G-G-B noisy raw image, where the aberration-degraded image $I_A$ in the sRGB domain is obtained. The ISP pipeline can be defined as:
\begin{equation}
\label{eq:gaoyao3}
    I_A = P_{GC} \circ P_{CCM} \circ P_{WB} \circ P_{demosaic} \circ (P_{mosaic}(I'_A) + N),
\end{equation}
where $N$ represents the Gaussian shot and read noise. $P_{mosaic}$ and $P_{demosaic}$ represent the procedures of mosaicking and demosaicking respectively.

To obtain accurate $PSF(x',y')$, we build a ray-tracing-based degradation model. For a spherical optical lens, its structure is determined by the curvatures of the spherical interfaces $c$, glass and air spacings $s$, and the refractive index $n$ and Abbe number $v$ of the material. Specifically, $n$ represents the refractive index at the ``$d$'' Fraunhofer line ($587.6nm$). Following~\cite{sun2021end}, we use the approximate dispersion model $n(\lambda) \approx A + B/\lambda^2$ to retrieve the refractive index at any wavelength $\lambda$, where $A$ and $B$ follow the definition of the ``$d$''-line refractive index and Abbe number. Thus, the lens parameters can be denoted as $\phi_{lens} = (c, s, n, v)$. Assuming that there is no vignetting, after the maximum field of view $\theta_{max}$ and the size of aperture stop $r_{aper}$ are determined, the ray tracing is performed. The traditional spherical surface can be expressed as:
\begin{equation}
\label{eq:gaoyao4}
    z = \frac{cr^2}{1+\sqrt{1-c^2r^2}},
\end{equation}
where $r$ indicates the distance from $(x, y)$ to the z-axis: $r^2=x^2+y^2$. Then, we conduct sampling on the entrance pupil. The obtained point $\textbf{S} = (x, y, z)$ can be regarded as a monochromatic coherent light source, and its propagation direction is determined by the normalized direction vector $\textbf{D} = (X, Y, Z)$. The propagation process of light between two surfaces can be defined as:
\begin{equation}
\label{eq:gaoyao5}
    \textbf{S}'=\textbf{S}+t\textbf{D},
\end{equation}
where $t$ denotes the distance traveled by the ray. Therefore, the process of ray tracing can be simplified as solving the intersection point $\textbf{S}'$ of the ray and the surface, together with the direction vector $\textbf{D}'$ after refraction. By building the simultaneous equations of Eq.~\ref{eq:gaoyao4} and Eq.~\ref{eq:gaoyao5}, the solution $t$ can be acquired. After substituting $t$ into Eq.~\ref{eq:gaoyao5}, the intersection point $\textbf{S}'$ can be obtained, while the refracted direction vector $\textbf{D}'$ can be computed by Snell's law:
\begin{equation}
\label{eq:gaoyao6}
    \textbf{D}' = \frac{n_1}{n_2} \bigg[ \textbf{D} + \bigg( \cos \langle{\textbf{p}, \textbf{D}}\rangle - \sqrt{\frac{n^2_2}{n^2_1}-1+\cos^2\langle{\textbf{p},\textbf{D}}\rangle} \bigg) \textbf{p} \bigg],
\end{equation}
where $\textbf{p}$ is the normal unit vector of the surface equation, $n_1$ and $n_2$ are the refractive indices on both sides of the surface, $\textbf{D}$ is the direction vector of the incident light, and $\cos \langle \cdot, \cdot \rangle$ is the operation for calculating cosine value between two vectors.
By alternately calculating the intersection point $\textbf{S}'$ and the refracted direction vector $\textbf{D}'$, rays can be traced to the image plane to obtain the PSFs. Under dominant geometrical aberrations, diffraction can be safely ignored (for the applied MOS with severe aberrations) and the PSFs can be computed through the Gaussianization of the intersection of the ray and image plane~\cite{li2021end}. Specifically, when the ray intersects the image plane, we get its intensity distribution instead of an intersection point. The intensity distribution of the ray on the image plane can also be described by the Gaussian function:
\begin{equation}
\label{eq:gaoyao7}
    E(m,n)=\frac{1}{\sqrt{2\pi}\sigma}\exp(-\frac{r(m,n)^2}{2\sigma^2}).
\end{equation}
$r(m,n)$ is the distance between the pixel indexed as $(m,n)$ and the center of the ray on the image plane, which is just the intersection point in conventional ray tracing, and $\sigma=\sqrt{\Delta x^2 + \Delta y^2}/3$. By superimposing each Gaussian spot, the final PSFs can be obtained. 

To sum up, the PSFs of all FoVs can be formulated as:
\begin{equation}
\label{eq:gaoyao8}
    PSF(x', y') = F(c, s, n, v, \theta_{max}, r_{aper}; x', y'),
\end{equation}
where $F(\cdot)$ refers to the setup ray tracing model. Therefore, we can produce the synthetic aberrated image by feeding the lens parameters $(c, s, n, v, \theta_{max}, r_{aper})$ to the imaging simulation pipeline.

\section{More Visual Results}
\label{sec:results}
\begin{figure*}[!t]
  \centering
  \includegraphics[width=0.8\linewidth]{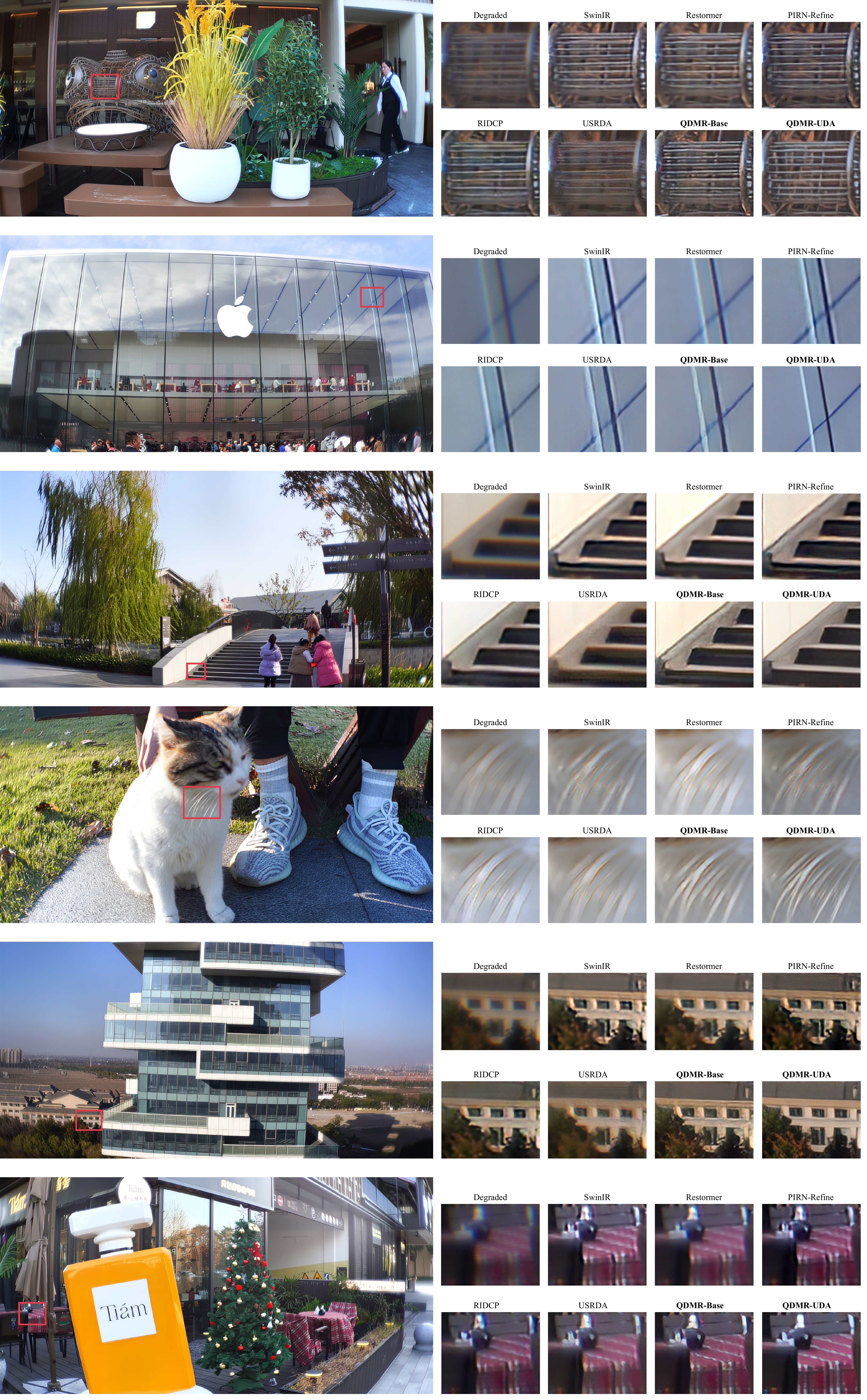}
  \caption{\textbf{More visual results of MOS-S1 on \textit{Real-Snap}.} Zoom in for the best view.}
  \vskip-4ex
  \label{fig:s1_real_world}
\end{figure*}

\begin{figure*}[!t]
  \centering
  \includegraphics[width=0.8\linewidth]{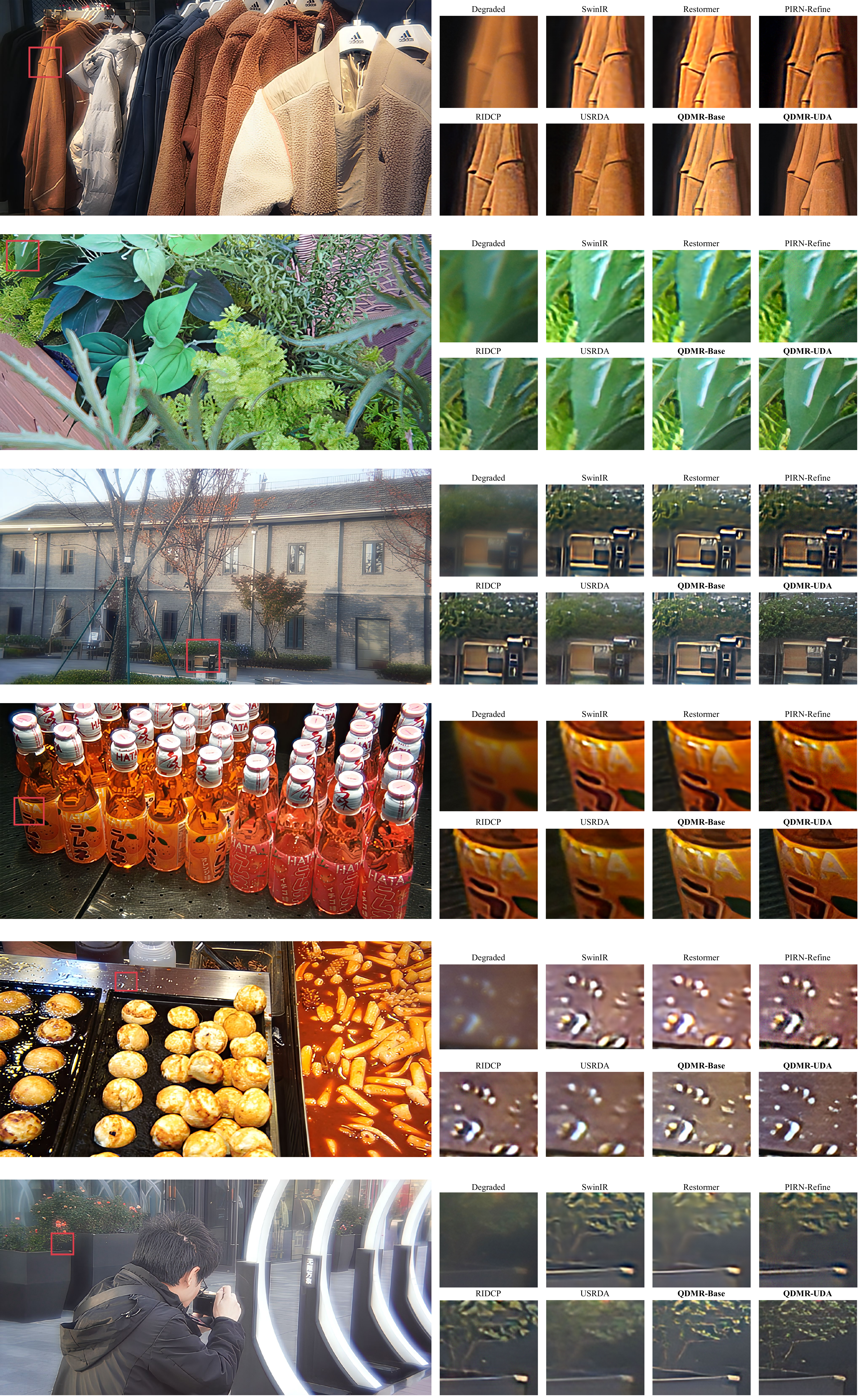}
  \caption{\textbf{More visual results of MOS-S2 on \textit{Real-Snap}.} Zoom in for the best view.}
  \vskip-4ex
  \label{fig:s2_real_world}
\end{figure*}

To better illustrate the effectiveness of the proposed QDMR in the DACAC task, we provide more visual results on \textit{Real-Snap} in Figures~\ref{fig:s1_real_world} and Figure~\ref{fig:s2_real_world}.
Compared to other methods, QDMR-UDA can provide high-quality CAC results with not only fewer artifacts but also richer and more realistic details and textures.
We also find that the results of the QDMR-Base model are rich in detail, but with non-negligible artifacts and too high contrast, which is largely due to the source-only training mode in the restoration stage.
The UDA strategies prove to be significant for unlocking the potential of QDMR in the restoration stage. 

